\documentclass[11pt]{article}

\usepackage[square,numbers,sort&compress]{natbib}
\usepackage{setspace}
\usepackage[margin=1in]{geometry}

\usepackage[T1]{fontenc}    
\usepackage[colorlinks=true, allcolors=blue]{hyperref}
\usepackage{hyperref}
\usepackage{url}            
\usepackage{booktabs}       
\usepackage{amsfonts}       
\usepackage{amsmath}
\usepackage{amssymb}
\usepackage{amsthm}
\usepackage{nicefrac}       
\usepackage{microtype}      
\usepackage{xcolor}         
\usepackage{enumitem}
\usepackage{bbold}
\usepackage{graphicx}

\usepackage{booktabs}
\usepackage{caption}
\usepackage{subcaption}

\theoremstyle{plain}
\newtheorem{definition}{Definition}
\newtheorem{assumption}{Assumption}
\newtheorem{remark}{Remark}
\newtheorem{theorem}{Theorem}
\newtheorem{proposition}[theorem]{Proposition}

\newtheorem{corollary}[theorem]{Corollary}

\newcommand{\KL}{\mathrm{KL}}

\newcommand{\E}{\mathbb{E}}

\newcommand{\R}{\mathbb{R}}
\newcommand{\Prob}{\mathbb{P}}

\newcommand{\calX}{\mathcal{X}}

\newcommand{\calC}{\mathcal{C}}
\newcommand{\calV}{\mathcal{V}}
\newcommand{\calF}{\mathcal{F}}
\newcommand{\calT}{\mathcal{T}}

\newcommand{\Rstar}{R^{*}}

\title{Limits of Reliability and Scaling in Language Models}

\author{%
  Subhabrata Majumdar
    \\
  Indian Institute of Management Bangalore\\
  Bengaluru, India \\
  \texttt{smajumdar@iimb.ac.in} \\
}
\date{}

\begin{document}

\maketitle

\begin{abstract}
Large language models (LLMs) are trained and evaluated as though perfect reliability is achievable for any task given sufficient scale. We show that this assumption is information-theoretically unjustified. Every generative task has a \textit{reliability ceiling} that no model can exceed, determined by how much output uncertainty is resolvable from observable context. The gap decomposes into a resolvable component closable with additional context and a subjective component inherent to task ambiguity. Autoregressive generation further degrades this ceiling at a rate governed by the task's \textit{dependency kernel}, which quantifies inter-token correlations in the output. From these two primitives, we derive a first-principles scaling law where LLM performance is bottlenecked by the scarcer resource: training data or model capacity. This law recovers the Chinchilla scaling law as a special case and provides a structural account of when scaling improves reliability. Beyond scaling, our framework unifies diverse practical phenomena, such as the benefits of retrieval-augmentation and the spectral mechanics of catastrophic forgetting. Our work formalizes the resource-complexity tradeoffs that govern model performance across domains, offering a unified theory of performance limits in generative language models.
\end{abstract}


\section{Introduction}
\label{sec:intro}

Large language models (LLMs) have demonstrated remarkable capabilities across a wide range of tasks, from code generation to mathematical proofs to creative writing. Yet, practitioners across the industry have observed a persistent empirical pattern that challenges the narrative of uniform progress: performance of LLMs on different tasks saturates at different levels, and the saturation points do not converge toward perfect reliability as models get larger~\citep{liang2023, grattafiori2024, akhtar2026aibenchmarksplateausystematic}. The power-law relationships that govern this improvement, known as the Chinchilla scaling law~\citep{kaplan2020, hoffmann2022}, empirically predict that cross-entropy loss decreases as a power of model size and training data, converging to an irreducible floor. Even though these relationships have become the primary tool for planning LLM training runs, they neither have exact theoretical interpretations, nor explain why frontier models achieve near-perfect accuracy on code generation and grade-school mathematics while doing much worse on creative writing, domain-specific reasoning, and multi-step agentic workflows \citep{raj2026consistencytestablepropertystatistical}.

This paper introduces the necessary theoretical framework to address these issues. We show that an LLM faces two distinct barriers to achieving perfect performance across generative tasks, both intrinsic to the task mixture which its training data comes from.

The first barrier pertains to \textit{reliability}. Every task has a maximum achievable reliability that no model can exceed, regardless of scale or architecture. This is determined by the conditional entropy in the output given the input. Verifiable tasks (such as formal proofs and code generation) have higher ceilings compared to more ambiguous tasks (such as creative writing). The reliability ceiling is further degraded by autoregressive generation, which compounds errors at a rate governed by the inter-token correlation structure of the task outputs. Injecting additional context during generation, such as through few-shot examples, retrieval, and tool use, slows down this degradation and raises the effective ceiling. 

The second barrier pertains to \textit{scaling}, and determines the rate at which scaling training data and model parameters reduce the gap to the ceiling. The derivation of scaling exponents \textit{a priori} has recently been identified as a central open problem in the emerging theory of deep learning~\citep{simon2026scientifictheorydeeplearning}. Based on the spectral structure of the task mixture, we derive the following power-law relationship between the empirical loss $\mathcal L(N,D)$, model size $N$, and training tokens $D$:
\begin{align}
    \mathcal{L}(N,D) &= \max \Bigl( \underbrace{\frac{A}{N^{(\bar{\mu}-1)/(d(\nu_{\calT}+1))}}}_{\text{approximation error}},\; \underbrace{\frac{B}{D^{(\bar{\mu}-1)/\nu_{\calT}}}}_{\text{estimation error}}\Bigr) \notag\\
    &  \quad + \underbrace{H(Y \mid X)}_{\text{irreducible error}} + \quad \text{lower order}.
    \label{eq:decomp_intro}
\end{align}
The exponents are determined by two spectral properties of the task mixture: how quickly the shared correlation structure across tasks decays ($\nu_{\calT}$) and how concentrated the prediction signal is in the leading modes of this structure ($\bar{\mu}$). The factor $d$ captures architectural overhead. The irreducible floor $H(Y|X)$ is the conditional entropy of the output given the input. This uncertainty is due to information relevant to generation that is absent from the input, persisting even with a perfect model. The $\max$ form indicates that performance is limited by whichever resource, data or capacity, is more scarce. The additive Chinchilla law \citep{hoffmann2022} emerges as a special case along the compute-optimal frontier where both resources are balanced.

\subsection{Setup and Our Contributions}
\label{sec:setup}
We specify a generative task by a joint distribution $P_{X,Y,C}$ over the input prompt $X$, the target output $Y = (y_1, \ldots, y_L)$ which is generated as a sequence of tokens from a finite vocabulary $\calV$, and $C$, the latent context containing information relevant to $Y$ but absent from $X$. A generative model, such as an LLM, aims to produce outputs that are acceptable proxies of the target output.

\begin{definition}
\label{def:gen_model}
A \emph{generative model} is a parametric family $Q_\theta$ that, given an input $X$, produces a distribution over $\calV^L$. The generated output $\hat{Y} = (\hat{y}_1, \ldots, \hat{y}_L)$ is a draw from this distribution.
\end{definition}

Transformers, the dominant architecture for current generative models, generates $\hat{Y}$ in an autoregressive manner. Formally, $Q_\theta(\cdot \mid X)$
factorizes as
$$
Q_\theta(\hat{y}_1, \dots, \hat{y}_L \mid X)
  = \prod_{t=1}^{L} Q_\theta(\hat{y}_t \mid \hat{y}_{<t}, X),
$$
where each factor is a regular conditional distribution on $\mathcal{V}$ given the sequence of already generated tokens $\hat y_{<t} = (\hat y_1, \ldots, \hat y_{t-1})$. Generation draws $\hat{y}_t \sim Q_\theta(\cdot \mid \hat{y}_{<t}, X)$ sequentially, and the sampling randomness at each step is independent
of $(Y, C)$ given $X$.

For a firm considering the deployment of LLMs in a number of generative tasks, the following questions are of interest. We answer them in this paper.

\begin{enumerate}[label={Q\arabic*.}]
    \item What is the performance limit of an LLM for a task the firm may automate, and how does this limit vary across tasks?
    \item When the firm runs a model in production for a task, how far below this limit does generated output fall, and how does this shortfall grow for longer outputs?
    \item What does it cost in training data and model capacity to take a model close to these limits, and what is lost when the model is specialized to one task?
\end{enumerate}

To answer Q1, we define the reliability ceiling ($\Rstar \in [0,1]$) as the fraction of output uncertainty resolvable from the observable input. Using the data processing inequality, we show that no generative model can produce task performance that exceeds $\Rstar$ (Theorem~\ref{thm:ceiling}). This bound is architecture-free and scale-free, requiring only the Markov structure induced by the model's access to the input. The gap to perfect reliability admits an economically and operationally meaningful decomposition,
$$ 1 - \Rstar = \delta_r + \delta_u,$$
where \(\delta_r\) is resolvable uncertainty that can, in principle, be eliminated by supplying additional relevant information, while \(\delta_u\) is unresolvable uncertainty arising from task ambiguity. This yields three regimes. \textbf{Fully verifiable} tasks have \(\delta_r=\delta_u=0\); \textbf{ideally verifiable} tasks have \(\delta_r>0,\delta_u=0\), so additional context can in principle eliminate the gap; and \textbf{unverifiable} tasks have \(\delta_u>0\), so a permanent reliability gap remains. In practice, a number of methods can help LLMs trained for generic tasks perform well in custom enterprise-relevant tasks by providing additional internal context. For example, retrieval-augmented generation \citep[RAG]{lewisetal2020_rag} fetches context most relevant to the input from an external database, few-shot prompting improves the performance of an LLM on a decisioning task (e.g. sentiment analysis) by providing examples of input-decision pairs, and tool use lets the model query external systems (e.g. calculator, code interpreter) whose outputs enter the context at generation time. We show that this happens because additional context from such methods raise the reliability ceiling of a task by closing its resolvable gap $\delta_r$ (Theorem~\ref{thm:context}).

To answer Q2, we introduce two quantities. The first quantity is sequence fidelity ($F_\theta$), which is the probability that the model-generated output falls within a task-appropriate tolerance of the target output. We establish two results for $F_\theta$: an impossibility result (Proposition~\ref{prop:fidelity-bridge}) stating that whenever the
output entropy, net of the redundancy the tolerance affords, exceeds
the information the input can supply, sequence fidelity is bounded
away from one for every model, and a large deviation bound (Proposition~\ref{prop:amplification}) which quantifies $F_\theta$ in terms of the ambiguity of the task and the accumulated generation errors. The second quantity dependency kernel ($K_Y$) is the matrix of conditional mutual information between output positions. Its shape is banded for tasks with short range dependencies and structured outputs (e.g. code generation), and dense for tasks with long range dependencies and free-form outputs (e.g. content creation). Proposition~\ref{prop:dep} shows that the accumulated generation error is bounded above in the banded case and bounded below in the dense case. Proposition~\ref{prop:verify} shows that this error can be minimized using the output of an oracle verifier that can intermittently check generated token sequences and rectify generation errors. 

The loss curves that guide pretraining budget decisions across large companies, and the finetuning tradeoffs that shape custom deployment of LLMs, are usually treated as empirical observations and regularities that are measured, fitted, and extrapolated with no account of where they come from. We derive both from our proposed framework to answer Q3. Under power-law assumption on the eigenspectrum of the dependency kernel of the training task mixture, Theorems~\ref{thm:beta} and~\ref{thm:alpha} identify the data and capacity exponents in Eq.~\eqref{eq:decomp_intro}, and Theorem~\ref{thm:decomp} assembles them with
the irreducible term $H(Y \mid X)$ into an overall power-law relationship in which loss is governed by the scarcer of the two resources. The same spectral
argument accounts for finetuning: Proposition~\ref{prop:finetune} shows that finetuning using additional data on single task reallocates representational capacity of the model across eigenmodes, with the performance loss on another task scaling 
as an inverse function of the alignment between the dependency kernels of the tasks.

\subsection{Related work}

\paragraph{Information bottleneck (IB) theory.}  The IB framework~\citep{tishby1999} formalizes the tradeoff between compression and prediction: a representation $T$ of input $X$ should minimize their mutual information (MI) $I(X;T)$ while maximizing $I(T;Y)$, the MI between the representation and output. \citet{shwartz2017} conjectured that deep networks implicitly 
optimize this objective, but subsequent work showed the compression 
phase depends on activation function choice~\citep{saxe2019} and 
can be mimicked by estimation artifacts~\citep{goldfeld2019}. 
Recent papers extend IB to generative 
models~\citep{lei2025revisitingllmreasoninginformation,
yang2025exploringinformationprocessinglarge}, but use it to 
analyze training dynamics rather than derive hard performance 
limits. We take the latter route: our reliability ceiling follows 
from the data processing inequality applied to the same Markov 
structure that motivates IB, but yields an upper bound on 
achievable performance rather than a characterization of what 
networks learn to discard.

\paragraph{LLM reliability and hallucination.}  
\citet{mohsin2025} prove via diagonalization that hallucination 
is inevitable for any computably enumerable model class, 
and~\citet{xu2024} establish similar impossibility results from 
a learning-theoretic perspective. These are existence proofs: they 
show failure must occur but do not quantify how much failure to 
expect on a given task. \citet{wang2020} formalize the 
distributional drift that drives compounding errors in 
autoregressive generation, and HELM~\citep{liang2023} documents empirically that performance saturation levels differ by task type. In contrast, our bounds are prescriptive. They formalize the reliability ceiling for a given task, decompose it into a closable 
component and a permanent floor, and identify interventions to close the first gap.

\paragraph{Neural scaling laws.}  
It is empirically established that language model loss follows a power law in model size $N$ and training tokens $D$ \citep{kaplan2020,hoffmann2022,besiroglu2024}. On the theoretical side, \citet{canatar2021} and \citet{bordelon2020} showed that 
power-law eigenspectra in the data covariance produce power-law 
learning curves in kernel regression, and multiple studies \citep{bahri2021,cagnetta2025learningcurvestheoryhierarchically,havrilla2024understandingscalinglawsstatistical} 
offered a statistical mechanics perspective. These results 
explain the functional form but leave the three terms of the 
Chinchilla law without task-theoretic interpretations. Recently, \citet{cagnetta2026derivingneuralscalinglaws} theoretically derived the data exponent in the Chinchilla law. We take a different route than them, and additionally yield the capacity exponent, the irreducible floor, as well as the $\max$-form linking the three as the fundamental relationship that governs learning over a task mixture. 

\paragraph{Task structure and learnability.}  
Classical PAC learning~\citep{valiant1984} bounds sample complexity 
via hypothesis class complexity but does not address the internal 
structure of the target. Research on benign 
overfitting~\citep{bartlett2020} and double 
descent~\citep{belkin2019} showed that the spectral structure of 
the data covariance governs generalization in overparameterized 
regimes. \citet{nayak2025information} model task structure through 
a bipartite skill-text graph, deriving Chinchilla-type scaling via 
an iterative concept-acquisition process. Our dependency kernel 
provides a complementary characterization: rather than modeling 
which skills a task requires, it captures how the output tokens of 
a task depend on one another, and how this correlation structure 
determines both autoregressive error propagation and scaling 
exponents.

\subsection{Outline}
The rest of the paper is organized as follows. Section 2 defines the reliability ceiling, establishes the reliability ceiling bound (Theorem~\ref{thm:ceiling}), develops the task taxonomy, and quantifies the effect of auxiliary context (Theorm~\ref{thm:context}). ection~\ref{sec:autoregressive} analyzes the degradation of sequence fidelity under autoregressive generation. Section~\ref{sec:dep_kernel} introduces the dependency kernel and derives its
consequences for error propagation and verification. Section~\ref{sec:training} derives the scaling law of Eq.~\eqref{eq:decomp_intro} and the spectral account of fine-tuning. Section~\ref{sec:discussion} provides a concluding discussion.

Proofs of all results are given in Appendix~\ref{app:proofs}. Two numerical illustrations accompany the theory, with experimental details in Appendix~\ref{app:exp}.

\section{The Reliability Ceiling}
\label{sec:formalism}

\subsection{Preliminaries}
A \emph{generative task} is specified by a joint distribution $P_{X,Y,C} \in \Prob$ over three random elements:
\begin{itemize}[nolistsep,leftmargin=*]
    \item $X \in \calX$: the \emph{observable context} (prompt, problem statement, or input),
    \item $Y = (y_1, \ldots, y_L) \in \calV^L$: the \emph{target output}, a sequence of tokens from a finite vocabulary $\calV$,
    \item $C \in \calC$: the \emph{latent context} (information relevant to $Y$ but absent from $X$).
\end{itemize}
We assume $(\Omega, \calF, \Prob)$ to be a probability space, and $\calX$ and $\calC$ to be Polish spaces with regular conditional distributions $P_{Y|X}$ and $P_{Y|X,C}$ respectively.  The variable $C$ encodes all task-relevant information not in $X$.  Its interpretation varies by task.  For formal mathematics or coding problems, $C = \emptyset$, since the problem statement fully specifies the answer. For factual question-answering, $C$ captures ambiguity and missing world knowledge, making $H(Y|X)$ nonzero.  For creative writing, $C$ encodes aesthetic preferences, cultural norms, and emotional context of both the writer and the reader, making $H(Y|X)$ large relative to $H(Y)$.

For a generative model with output $\hat Y$ (Definition~\ref{def:gen_model}), the generation process factors as $X \to T \to \hat{Y}$, where $T$ is the model's internal representation of $X$. Since the model observes only $X$, never $C$, $T$ and $C$ are conditionally independent given $X$, written $T \perp C \mid X$.  Similarly, the generated output $\hat{Y}$ and the input $X$ are conditionally independent given $T$, written $\hat{Y} \perp X \mid T$, as the decoder conditions on $T$ alone\footnote{Modern architectures use residual connections or cross-attention that let the decoder access $X$ alongside $T$.  This does not violate the condition: whatever the decoder sees is a function of $X$ and can be absorbed into a richer definition of $T$.}. These two independence conditions, together with the fact that $X$ is a marginal of $(X,C)$, yield the Markov chain
\begin{equation}
    Y \longleftrightarrow (X, C) \longleftrightarrow X \longleftrightarrow T \longleftrightarrow \hat{Y},
    \label{eq:markov}
\end{equation}
where each link discards information: $(X,C) \to X$ loses latent context, $X \to T$ loses what the encoder discards, and $T \to \hat{Y}$ loses what the decoder does not use.

\begin{definition}
\label{def:reliability-ceiling}
The \emph{reliability ceiling} of a generative task is $\Rstar :=I(X;Y)/H(Y) \in [0,1]$.
\end{definition}

The quantity $\Rstar$ measures the fraction of output uncertainty resolvable from the observable input.  When $\Rstar = 1$, the task is fully determined by $X$.  When $\Rstar < 1$, a nonzero fraction depends on latent context $C$ that no model can access.  The decomposition $I(X;Y) = H(Y) - H(Y|X)$ clarifies that $\Rstar$ is a property of the task distribution, not of any particular model.

\begin{theorem}
\label{thm:ceiling}
Let $P_{X,Y,C}$ define a generative task and $Q_\theta$ a generative model.  Then for any measurable functional $\phi: \calV^L \to \R$,
\begin{equation}
    I(\phi(\hat{Y});Y) \;\leq\; I(T;Y) \;\leq\; I(X;Y).
    \label{eq:ceiling_chain}
\end{equation}
Consequently, no model can achieve reliability exceeding $\Rstar$: $\;\sup_{Q_{\theta}} I(\hat{Y};Y)/H(Y) \leq \Rstar$.
\end{theorem}

The inequality chain~\eqref{eq:ceiling_chain} follows from applying the data processing inequality to each link of the Markov chain~\eqref{eq:markov}: information can only be lost, never created, as we move from the joint $(X,C)$ through the encoder to $T$ and through the decoder to $\hat{Y}$. The bound is \emph{architecture-free} (it holds for transformers, diffusion models, or any parametric family) and \emph{scale-free} (increasing parameters, data, or compute cannot overcome it).  It also applies to any functional $\phi$ of the output: not just the raw token sequence but any downstream quantity computed from it, such as an extracted answer, a summary, or a classification label.

\subsection{Task taxonomy}
The gap $1 - \Rstar$ admits a further decomposition that clarifies what kinds of latent context drive it.  We write $C \equiv (C_r, C_u)$, where $C_r$ is \emph{resolvable} context: information that exists and could in principle be provided to the model (a relevant codebase, the user's style preferences, factual knowledge retrievable from a database). $C_u$ is \emph{unresolvable} context: subjective information that has no fixed value because the ``correct'' output depends on the reader, the cultural moment, or aesthetic judgments that cannot be specified even in principle.  By the chain rule:
\begin{equation*}
    1 - \Rstar \;=\; \underbrace{\frac{I(C_r;\, Y \mid X)}{H(Y)}}_{\delta_r:\;\text{resolvable gap}} \;+\; \underbrace{\frac{I(C_u;\, Y \mid X, C_r)}{H(Y)}}_{\delta_u:\;\text{subjective gap}}.
\end{equation*}
The resolvable gap $\delta_r$ can be closed by enriching the input, e.g. by better prompting, retrieval, or tool use, effectively raising the reliability ceiling to $\Rstar_{\max} := 1 - \delta_u$ (Section~\ref{sec:context}). The subjective gap $\delta_u$ cannot be closed by any intervention, because the information it represents does not have a determinate value. This formalization relates to prior work on uncertainty decomposition in LLMs \citep{Lietal2023cue,walha2025finegraineduncertaintydecompositionlarge}, and resolvable vs. irresolvable disagreement in crowdsourced data annotation \citep{schaekermann2018resolvable}. 

The above decomposition yields a task taxonomy based on the structural character of the gap:
\begin{itemize}[nolistsep,leftmargin=*]
    \item \textbf{Fully verifiable} ($\delta_r = \delta_u = 0$, so $\Rstar = 1$): The output is a deterministic function of the input.  Examples: formal proofs, arithmetic, code with complete specifications.
    \item \textbf{Ideally verifiable} ($\delta_u = 0$, $\delta_r > 0$, so $\Rstar < 1$ but $\Rstar_{\max} = 1$): The output is deterministic given all relevant context, but some context is absent from the input.  The gap is entirely closable in principle.  Examples: factual QA (retrieval can help), code with ambiguous specs (clarification can help), medical diagnosis (more tests can help).
    \item \textbf{Unverifiable} ($\delta_u > 0$, so $\Rstar_{\max} < 1$): No amount of context can make the output deterministic, because correctness depends on subjective factors with no fixed value.  Examples: creative writing, open-ended advice, culturally situated rhetoric.
\end{itemize}


This taxonomy determines a task's fundamental suitability for reliable generative modeling: fully and ideally verifiable tasks can in principle reach $\Rstar = 1$; unverifiable tasks face a permanent floor $\delta_u$ that no model, architecture, or input engineering can overcome\footnote{For a parallel complexity-theoretic interpretation, see Appendix~\ref{app:complexity}.}. Note that the classification above refers to functional verifiability, not token-level determinism.  For code generation, many distinct token sequences (differing in variable names, formatting, or algorithmic approach) satisfy the same specification, so $H(Y \mid X) > 0$ when $Y$ is a raw token sequence. But the reliability ceiling $R^* = 1$ would hold with respect to functional equivalence, i.e. the set of outputs that pass a test suite. We formalize this task-appropriate notion of equivalence later in Assumption~\ref{ass:regularity}.

\paragraph{Numerical Illustration 1}(Performance ceilings in LLM benchmarks).
To find evidence of differential reliability ceilings across task types, we adapt the benchmark saturation framework of \citet{akhtar2026aibenchmarksplateausystematic}. They define the saturation index of a benchmark dataset as $S_\text{index} := \exp(-R_\text{norm}^2)$, where $R_\text{norm}$ is the scaled difference between the first and $k^\text{th}$ highest scores of LLMs on that benchmark. High $S_\text{index}$ indicates strong evidence of model-level performance saturation, with task performance hitting an empirical ceiling at the highest score. 

\begin{figure}[ht]
    \centering
    \includegraphics[width=\linewidth]{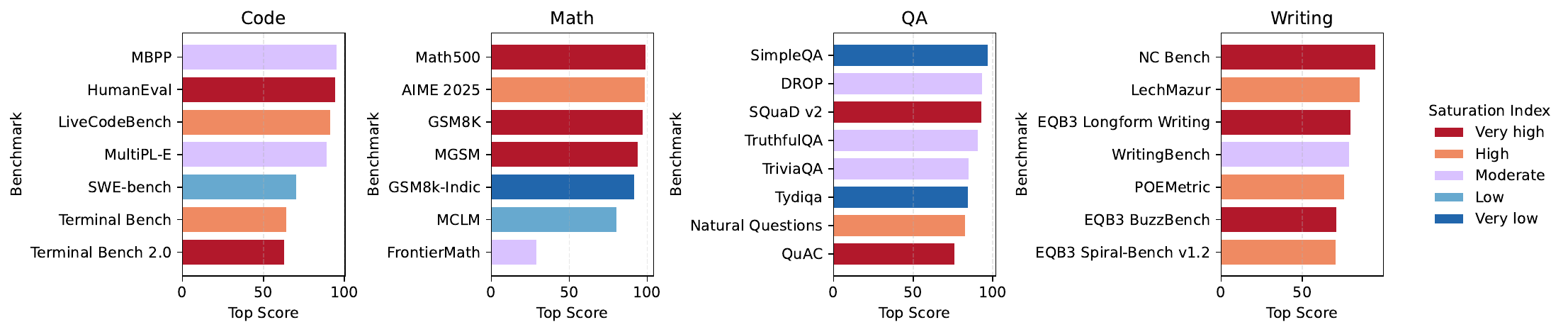}
    \caption{Best benchmark scores by task type. Bins of $S_\text{index}$, based on $k=5$ top scores, are Very low ($< 0.01$), Low ($[0.01, 0.3)$), Moderate ($[0.3, 0.7)$), High ($[0.7, 0.9)$), and Very high ($\geq 0.9$).}
    \label{fig:reliability_ceiling}
\end{figure}

Figure~\ref{fig:reliability_ceiling} shows the best score by an LLM and saturation index for benchmarks across four types of tasks: code, math, Q\&A, and writing (details in Appendix~\ref{app:benchmark_sat}). Most writing benchmarks have top scores well below the maximum, with all but one showing high saturation: suggesting the presence of unresolvable context. On the other hand, math benchmarks (with exact verifiers) that show high saturation all have near-perfect top scores. Q\&A benchmarks split revealingly. High-scoring Q\&A benchmarks tackling simpler tasks (e.g. SimpleQA) show less saturation, suggesting that as models become more capable, they progressively tackle resolvable context better. However, more ambiguous Q\&A benchmarks (Natural Questions, QuAC) show high saturation at well below perfect score.

\subsection{Reliability celing in practice}
\label{sec:context}
While the ceiling is a property of the task, it constrains two things practitioners control. The first is how much context to supply. Every technique that enriches the input can potentially raise the reliability ceiling to the extent the unresolvable gap $\delta_u$ allows. The second is the interpretation of a benchmark evaluation. Scoring against a single reference presumes the task determines its output. Where it does not, the resulting noise is a property of the task ambiguity rather than of the evaluation protocol.

\subsubsection{Effect of additional context}
\label{sec:context}

Techniques to improve the performance of LLM-based systems---such as Retrieval Augmented Generation (RAG), few-shot prompting, tool use, and source-grounded question-answering---all follow a similar strategy. Each makes a portion of the latent context $C_r$ observable, raising $\Rstar$ toward $\Rstar_{\max}$.  For ideally verifiable tasks, $\Rstar_{\max} = 1$ and the path to perfect reliability is an engineering problem.  The following result quantifies this gain and its limit.

\begin{theorem}
\label{thm:context}
Let $\tilde C$ be auxiliary context jointly distributed with $(X, Y,
C)$ and made available to the model. Assume $\tilde C \perp C_u \mid
(X, C_r)$, i.e., the auxiliary context is informative only through the
resolvable component of the latent context. Then:
\begin{enumerate}[label=(\roman*),nolistsep,leftmargin=*]
\item $\tilde \Rstar := I(X, \tilde C; Y)/H(Y) \ge \Rstar$, with
equality if and only if $\tilde C \perp Y \mid X$.
\item $\tilde \Rstar - \Rstar = I(\tilde C; Y \mid X)/H(Y) \le
\delta_r$, i.e.\ the improvement in reliability ceiling is bounded by
the resolvable gap.
\item For a sequence of auxiliary variables $\tilde C_1, \tilde C_2,
\ldots$, the augmented ceiling $\tilde \Rstar$ is nondecreasing in the
number of injections. When $\tilde C$ resolves all of $C_r$, the
ceiling reaches $\Rstar_{\max} = 1 - \delta_u$.
\end{enumerate}
\end{theorem}

Part (i) gives the general mechanism. Additional context makes the output more
determined, reducing the gap to perfect reliability. Part (ii)
establishes the limit: context enrichment can close the resolvable gap
$\delta_r$ but leaves the subjective gap $\delta_u$ untouched. Part
(iii) shows that successive injections have nondecreasing but bounded
returns. For ideally verifiable tasks ($\delta_u = 0$), sufficiently
rich context can in principle push $\Rstar$ to $\Rstar_{\max} = 1$, and
the path to perfect reliability is an engineering problem. For
unverifiable tasks ($\delta_u > 0$), a permanent floor on unreliability
remains regardless of how much context is supplied.

\subsubsection{Instability in benchmark evaluation}
Current evaluation practices implicitly assume that all tasks have $\Rstar = 1$: a model is scored against a single ground truth, and the resulting accuracy or reliability metric is treated as a reliable quality measure.  When $\Rstar < 1$, the task admits multiple legitimate outputs for the same input, and evaluation against any single reference introduces uncertainty that cannot be eliminated.

\begin{proposition}
\label{prop:misspec}
Consider a set of LLMs evaluated on a benchmark of $M$ instances drawn i.i.d.\ from a task with $R^* < 1$. Suppose each output is scored by exact match against a reference drawn from $P(Y \mid X)$ independently of the output, and
that every model matches the reference with positive probability on
each instance. Then the correlation between the benchmark scores of LLMs across
two independent evaluation runs is strictly less than one.
\end{proposition}

LLM leaderboards, such as OpenLLM v2\footnote{https://huggingface.co/collections/open-llm-leaderboard/open-llm-leaderboard-2}, regularly aggregate evaluation scores across multiple benchmarks. Proposition~\ref{prop:misspec} shows that such aggregate leaderboard scores are misleading when they combine tasks with different $\Rstar$ values, as strong performance in verifiable tasks ($R^*=1$) can potentially mask noisy rankings in unverifiable ($R^*<1$) tasks.
\section{Autoregressive degradation of output quality}
\label{sec:autoregressive}

Theorem~\ref{thm:ceiling} bounds what any model can achieve on a generative task, but it treats the output $\hat Y$ as a single atomic object.  In practice, autoregressive models generate $\hat{Y} = (\hat{y}_1, \ldots, \hat{y}_L)$ token by token, conditioning each step on the previously generated prefix $\hat{y}_{<t}$.  Once an error occurs at some step, all subsequent tokens are conditioned on a corrupted prefix, which the conditional distribution $P(\cdot \mid \hat y_{<t}, X)$ away from the true $P(\cdot \mid y_{<t}, X)$.  This compounding mechanism, known as exposure bias~\citep{wang2020}, is not captured by the single-shot ceiling of Theorem~\ref{thm:ceiling}. 

To that end, we ask two questions in this section: does the ceiling $R^*$ still constrain what a model can produce over a long
sequence? And how fast does the probability of producing an acceptable
sequence fall under autoregressive generation?

\subsection{Sequence fidelity}
Both questions need a notion of success for the whole sequence. Let $Y =
(y_1, \ldots, y_L)$ be the reference output and $\hat Y = (\hat y_1,
\ldots, \hat y_L)$ the model's output. Fix a distance metric $d$ between
sequences of equal length and a tolerance $\eta > 0$: so the output counts
as acceptable when $d(\hat Y, Y) \le \eta$. For code, $d$ can count
failed tests.The metric $d$ captures the appropriate notion of similarity for the task at hand, such as exact token match or passing a test suite for code, semantic similarity for prose, and entailment for factual QA. Formally, we assume an equivalence relation on prefixes, $y_{<t} \sim y'_{<t}$ when the two carry the same meaning, such that the task law depends on a prefix only through its class: $P(\cdot \mid y_{<t}, X) = P(\cdot \mid y'_{<t}, X)$ whenever $y_{<t} \sim y'_{<t}$.
The distance $d$ is then a metric on classes. For code the classes can
be singletons and $d$ an edit or test count.

\begin{assumption}
\label{ass:distance}
The distance metric $d$ has the following properties
\begin{enumerate}[label=(\alph*),leftmargin=*,nolistsep]
    \item \textbf{Monotonicity}: $d([u\,z], [v\,z']) \ge d(u, v)$ for all
    prefixes $u, v$ and tokens $z, z' \in \mathcal V$.
    \item \textbf{Extension contraction}: $d([u\,z], [v\,z]) \le d(u, v)$
    for prefixes $u, v$ and common token $z \in \mathcal V$.
\end{enumerate}
\end{assumption}

Monotonicity ensures that extending two sequences never brings them closer, so that a generation error in a previous token is not undone by later tokens. The second property posits that appending the same token to both prefixes never increases the distance between them. Formally, we assume an equivalence relation on prefixes,
$y_{<t} \sim y'_{<t}$ when the two carry the same meaning, such that
the task law depends on a prefix only through its class: $P(\cdot \mid
y_{<t}, X) = P(\cdot \mid y'_{<t}, X)$ whenever $y_{<t} \sim y'_{<t}$.

\begin{definition}
\label{def:seq-fidelity}
The \emph{sequence fidelity} of a model $Q_\theta$ is the probability that the model produces an output $\hat Y$, that is acceptable, with respect to the reference output $Y$, at a tolerance threshold $\eta \geq 0$:
$$ F(Q_\theta; \eta) \equiv F_\theta := \mathbb P\big[\, d(\hat Y, Y) \le \eta
\,\big].$$
\end{definition}

The following result connects the population-level reliability ceiling with the model-level sequence fidelity.

\begin{proposition}
\label{prop:fidelity-bridge}
Let $N_\eta := \sup_{\hat y \in \mathcal V^L} \big|\{ y \in \mathcal V^L
: d(\hat y, y) \le \eta \}\big|$ be the largest number of reference answers, i.e. the number of answers that count as correct at once, that fit within tolerance of any single output. Then, for every generator whose sampling randomness is independent of $Y$ given $X$,
$$ 1 - F_\theta \geq \frac{H(Y) - \log (2 N_\eta)}{L \log | \mathcal V |} - R^*.$$
\end{proposition}

The output contains $L \log|\mathcal V|$ bits. Specifying $Y$ requires
$H(Y)$ bits on average, but $\log( 2 N_\eta )$ of these are redundant for our purpose since it suffices to produce any one of the $N_\eta$ sequences within
tolerance of $Y$. The ratio on the right is therefore the number of bits the
model must supply, divided by the number of bits it emits. By
Theorem~\ref{thm:ceiling}, the input $X$ supplies at most a fraction
$R^*$ of them. The failure probability $ 1 - F_\theta $ is at least the difference
between these two fractions. This bound is informative when the difference is positive, that is, when $H(Y) - \log( 2 N_\eta ) > R^* L \log|\mathcal V|$. This holds for tasks with high entropy but a small tolerance. For near-deterministic tasks, $H(Y)$ is much smaller than $L \log|\mathcal V|$ and the bound is trivial.

\subsection{Degradation under autoregressive generation}
We now analyze the generation process for a further characterization of $F_\theta$. When the generator samples a token at step $t$ of the generation process: $\hat y_t \sim Q_\theta(\cdot \mid \hat y_{<t}, X)$, two things can go wrong. First, the model's conditional can differ from the true one by more than tolerance $\eta$ in the metric $d$. We measure this using \emph{model error}
$$
\epsilon_t := \Prob_{z \sim Q(\cdot \mid \hat y_{<t}, X)}
\left[ d([\hat{y}_{< t}\, z], y_{\leq t}) > \eta \mid \hat{y}_{<t} \right].
$$
Second, the prefix it conditions on can already be wrong, which we measure using \emph{drift at $t$}, defined as $r_t := d(\hat y_{<t}, y_{<t})$, the distance of the generated prefix from the reference prefix.

In order to quantify how these errors affect the fidelity of an output generated by the model, we make the following assumptions on the data distribution. We use $\mathrm{KL}(\cdot||\cdot)$ and $ H(\cdot||\cdot)$ to denote KL divergence and Hellinger distance, respectively.

\begin{assumption}
\label{ass:regularity}
The following hold for all $X$ and prefixes $y_{<t}, y'_{<t}$,

\begin{enumerate}[label=(\alph*),leftmargin=*,nolistsep]
\item \textbf{Regularity.} There is a nondecreasing $\omega_u :
[0,\infty) \to [0,\infty)$ with $\omega_u(0) = 0$ such that 
\begin{align}
\label{eq:regularity}
& \mathrm{KL}\big( P(\cdot \mid y_{<t}, X) \,\big\|\, P(\cdot \mid y'_{<t},
X) \big) \notag\\
&  \quad \leq \omega_u\big( d(y_{<t},\, y'_{<t}) \big).
\end{align}

\item \textbf{Sensitivity.} There is a nondecreasing $\omega_\ell :
[0,\infty) \to [0,1]$ with $\omega_\ell(0) = 0$ such that
\begin{align}
\label{eq:sensitivity}
H^2\big( P(\cdot \mid y_{<t}, X),\; P(\cdot \mid y'_{<t}, X) \big)
&\geq \omega_\ell\big( d(y_{<t},\, y'_{<t}) \big).
\end{align}

\end{enumerate}
\end{assumption}


We assume a well-behaved data generating distribution that is (a) continuous, so that nearby prefixes impose nearby next-token laws (Eq.~\ref{eq:regularity}), and (b) sensitive, so that the next-token law changes if two prefixes are too far apart (Eq.~\ref{eq:sensitivity}). 

We also assume that errors are additive in the metric $d$. A sequence is acceptable when its total error stays within the budget $\eta$. A run that has already drifted by $r_t$ has spent $r_t$ of that budget, so the tokens that keep it acceptable are those that fit in what is left.

\begin{assumption}
\label{ass:budget}
For $P$-almost every $(X,Y)$, every $t \le L$, prefix $u$ of
length $t-1$, and token $z \in \mathcal V$,
\[
d\big( [u\,z],\; y_{\le t} \big) \;\ge\; d\big( u,\; y_{<t} \big) \;+\;
d\big( [y_{<t}\,z],\; y_{\le t} \big).
\]
\end{assumption}

Errors already committed and the error committed now add up. Deviation
counts and unnormalized edit counts satisfy this with equality, and so
does any metric that scores a sequence by summing positional penalties.

\begin{proposition}
\label{prop:amplification}
Let Assumptions \ref{ass:distance}-\ref{ass:regularity} hold, and  $\epsilon_t$ be the model error and $r_t$ be the drift at step $t$. Let $\epsilon_t^* := \Prob_z[d([y_{< t}\, z], y_{\leq t}) > \eta \mid \hat y_{<t}]$ be the model error under the true prefix. Then

\begin{enumerate}[label=(\roman*),leftmargin=*,nolistsep]
    \item The following holds almost surely
    $$ \epsilon_t \geq \epsilon^*_t + g_t; \quad
    g_t := \Prob_z \big[ \, \eta - r_t < d( [y_{<t}\, z], y_{\le t} ) \le \eta \mid \hat y_{<t} \big].$$
    
    \item  Suppose $\epsilon^*_t \ge \epsilon > 0$ for every $t
    \le L$ whenever the output so far is within tolerance. Then for every $D \ge 0$ and $G_L := \sum_{t=1}^L g_t$,
    \begin{align*}
    \mathbb P\Big[\, d( \hat Y, Y ) \le \eta \;\text{ and }\;
    G_L \ge D \,\Big] &\le e^{-(\epsilon L + D)}.
    \end{align*}
    \item The failure probability satisfies
    $$ 1 - F_\theta \;=\; \mathbb P\big[\, d( \hat Y, Y ) > \eta
    \,\big] \;\ge\; 1 - \gamma(L) e^{-\epsilon L};\quad
    \gamma(L) := \left( \mathbb E e^{-G_L} \right)^{1/2}. $$
\end{enumerate}
\end{proposition}

Under exact token match ($d(\hat{y}, y) = \mathbb 1 [\hat{y} \neq y]$), part (i) of the proposition reduces to the statement
that any single token error is terminal: the drift becomes positive,
the generated sequence leaves the tolerance band, and by monotonicity
it never returns. Under a semantic metric, where $d(\hat y_{<t},
y_{<t}) = 0$ for paraphrases, synonym substitutions, and other
surface-level variations, the drift $r_t$ is zero whenever the
generated prefix remains semantically equivalent to the target, and the
amplification term vanishes with it. The drift becomes positive only
when the model has committed to a \emph{meaning} from which the target
cannot be recovered, which is the operationally relevant notion of
error for most natural language tasks.

Part (ii) and (iii) accumulate this over the sequence. We posit that the task is ambiguous at a resolution $\eta$ around the true answer and at every step of the generation process a token exceeds this budget on its own has probability at least $\epsilon$. Drift due to earlier errors \textit{adds to} the exponent in the form of $\gamma(L)$, with one unit of decay for each unit of accumulated
error. Both parts come from one inequality, $\mathbb E[ \mathbb 1_{B_L}
e^{G_L} ] \le e^{-\epsilon L}$ with $B_L$ the event that the output
stays within tolerance throughout (see Proof in Appendix~\ref{sec:proof_prop_amplification}). This is a Feynman-Kac bound for the generation process until the first tolerance violation. The weight $e^{G_L}$ accumulates along the outputs that have not yet failed, so $F_\theta$ is a survival probability and $\epsilon$ bounds its decay rate.

\begin{remark}
Depending on the task, it is possible to slow down the degradation
using intermittent oracle verifiers that reset the drift $r_t$. The
effectiveness of such verifiers depends on whether local constraints
are checkable locally using the dependency structure of past tokens
(see Section~\ref{sec:verification}).
\end{remark}

\section{Characterizing Tasks using Dependency Kernel}
\label{sec:dep_kernel}

Proposition~\ref{prop:amplification} shows that autoregressive generation degrades $F_\theta$ by a factor governed by the errors accumulated in the output as it drifts, but does not resolve \emph{what determines the drift}. Two tasks with identical output entropy $H(Y)$ and reliability ceiling $\Rstar$ can differ dramatically in how their tokens depend on one another. A code generation task and a poetry task may have the same $H(Y)$ and $\Rstar$, yet errors in code tend to be locally contained, whereas errors in poetry affect the quality of the entire generation.

\subsection{Dependency kernel}

We introduce the \emph{dependency kernel}\footnote{Our use of ``dependency kernel'' is information-theoretic, and unrelated to syntactic dependency kernels in NLP~\citep{bunescu2005shortest}.} to quantify the internal correlation structure of the output of a task and explain why drift propagation differs across task types.


\begin{definition}
\label{def:dep_kernel}
For a generative task with target $Y = (y_1, \ldots, y_L)$ and observable context $X$, the \emph{dependency kernel} is the $L \times L$ matrix
\begin{equation}
    K_Y(t,t') \;:=\; I(y_t;\, y_{t'} \mid y_{-\{t,t'\}},\, X), \quad 1 \leq t, t' \leq L,
    \label{eq:dep_kernel}
\end{equation}
where $y_{-\{t,t'\}}$ denotes all tokens except positions $t$ and $t'$.
\end{definition}

The entry $K_Y(t,t')$ measures how much knowing token $t'$ tells us about token $t$, after controlling for the rest of the sequence and the input.  
If tokens $t$ and $t'$ are correlated only because both depend on an intermediate token $s$, then $K_Y(t,t') = 0$ after conditioning on $y_{-\{t,t'\}}$
.

\begin{figure}[ht]
    \centering
    \includegraphics[width=\linewidth]{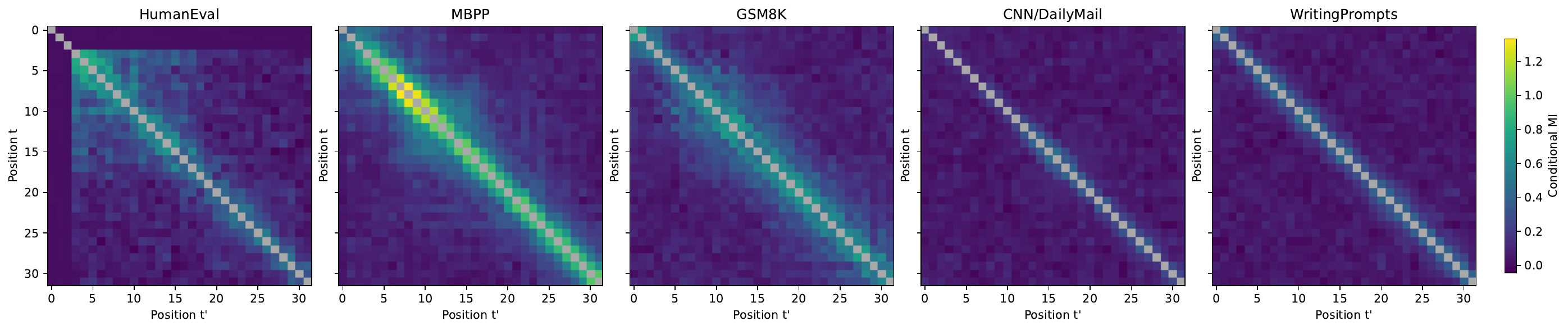}
    \caption{$I(y_t;y_{t'})$ for public benchmarks, with $L = 32$. Coding (HumanEval, MBPP) and math (GSM8K) tasks show banded or block diagonal structure, whereas ambiguous tasks (CNN/DailyMail and WritingPrompts) have dense kernels with smaller elements.}
    \label{fig:dependency_kernel}
\end{figure}

\paragraph{Numerical Illustration 2}(Mutual information in LLM benchmark outputs).
For fully verifiable tasks like code generation, $K_Y$ is approximately \emph{banded}. Elements near the diagonal reflect local syntactic constraints (matching brackets, type consistency, sequential statements), while sparse off-diagonal entries correspond to variable references, function calls, and import dependencies. 
For ideally verifiable tasks like scientific writing, $K_Y$ is approximately \emph{block-diagonal}. Within-paragraph dependencies are strong (each sentence supports the paragraph's claim), cross-paragraph links are mediated by topic coherence, and global constraints (thesis consistency, citation accuracy) create sparse long-range entries. 
For unverifiable tasks like creative writing, $K_Y$ is \emph{dense}. Rhyme schemes create dependencies between line endings, metaphorical coherence links semantically distant passages, and the tonal arc constrains word choice globally. 
As proxy of $K_Y(t,t')$, Figure~\ref{fig:dependency_kernel} plots the (unconditioned) MI $I(y_t; y_{t'})$ calculated on five benchmark datasets (see Appendix~\ref{app:dep_kernel} for implementation details). 

\subsection{Dependency-modulated degradation}

Proposition~\ref{prop:amplification} shows that autoregressive degradation is governed by drift $\gamma(L)$, but treats drift as a single aggregate quantity without decomposing it by source.  The dependency kernel provides this decomposition. An error at position $t_0$ contributes to the drift at position $t$ in proportion to $K_Y(t_0, t)$, that is, based on how strongly $t$ depends on $t_0$.

\begin{proposition}
\label{prop:dep}
Suppose that the drift amplification at a step is proportional to the kernel mass of the positions already deviated from: $g_t \asymp \sum_{s<t} K_Y(s,t)\, \xi_s$. Then, for constants $c_1, c_2 >0$,
\begin{enumerate}[label=(\roman*),nolistsep,leftmargin=*]
    \item For banded $K_Y$ with bandwidth $w$ (i.e., $K_Y(t_0,t) = 0$ for $|t - t_0| > w$): $\gamma(L) \geq \exp(-c_1 L)$.
    \item For dense $K_Y$ with off-diagonal mass bounded below: $K_Y(s,t) \ge \kappa/L$, $\gamma(L) \leq 2 \exp(-c_2 L)$.
\end{enumerate}
\end{proposition}

Thus, degradation beyond the task's own ambiguity is confined to a constant
factor per unit length in the banded case and forced to an exponential
in the dense case.
This is the formal expression of the intuition that code is more robust
to autoregressive errors than poetry. The banded dependency structure
of code confines error propagation to a local neighborhood, while the
dense structure of poetry allows a single early error to corrupt the
global coherence of the entire output.

\begin{remark}
The banded and dense cases differ in a way that is familiar from
interacting particle systems. Under a banded kernel the accumulated
amplification $G_L$ is an additive functional of a finite-range
interaction, and its large-deviation behaviour is governed by a
transfer operator with a spectral gap. Under a dense kernel, $G_L$ is quadratic in the empirical
density of deviations, which is the Curie-Weiss scaling of a
mean-field model. 
Establishing a large-deviation principle for $G_L$ under a
dense kernel, and determining whether the tilted measure undergoes a
transition, would sharpen the constants and give the banded-dense
contrast a sharper description than the one we prove here. We also note
that no bound of this kind can decay faster than exponentially in $L$,
since $g_t \le 1$ forces $G_L \le L$.
\end{remark}

\subsection{Verification slows degradation}
\label{sec:verification}

A natural mitigation of autoregressive degradation is \emph{verification}: an oracle $v_t = v(\hat y_{\le t}, X)$ that inspects the prefix and, when it finds a deviation, corrects it before generation proceeds.  Specifically, an oracle that resets the prefix at step $t$ results in the drift term $r_t = 0$, so the amplification probability $g_t$. becomes zero.

\begin{proposition}
\label{prop:verify}
Let $\hat Y^{\mathrm{ver}}$ be generated with an oracle that inspects
the prefix and, on detecting a deviation, replaces it by one at zero
drift, and suppose the oracle never increases the drift, so
$r_t^{\mathrm{ver}} \le r_t$ at every step. Then
\[
g_t^{\mathrm{ver}} \;\le\; g_t \quad \forall t, \qquad
\Rightarrow \qquad \gamma^{\mathrm{ver}}(L) \;\ge\; \gamma(L) .
\]
\end{proposition}

Without verification, a deviation early in a output raises the amplification at every position that follows it, so the accumulated $G_L$ grows with the length. With verification it raises the amplification only until the next
inspection.
 
Whether the oracle of the proposition exists is determined by the dependency structure. For banded $K_Y$,
as in code, correctness at position $t$ is largely determined by local
constraints, syntax, types, bracket matching, that a prefix-based
oracle can check. Deviations are detected close to where they occur,
so the premise of Proposition~\ref{prop:verify} is approximately met and the correction is substantial. For dense $K_Y$, as in poetry, correctness at position $t$ depends on the entire sequence: whether the word contributes to the
rhyme scheme, thematic arc, and tonal coherence of the whole. No
prefix-based oracle can assess these properties as they depend on tokens not yet generated. This explains why constrained decoding and
grammar-guided generation are effective for code, JSON, and SQL but
offer no benefit for creative writing.

\begin{remark}
Together with Theorem~\ref{thm:context}, these results reveal a triple
penalty for unverifiable tasks with dense dependencies: a depressed
ceiling $\Rstar$ from $\delta_u > 0$, fast decay of $\gamma(L)$ from
error propagation through the dense kernel, and ineffective
verification, since quality is a global property that no prefix-based
oracle can assess. Fully verifiable tasks with banded dependencies
enjoy the opposite advantage on all three fronts.  
\end{remark}

%

\section{Training Language Models at Scale}
\label{sec:training}
The results so far characterize what a trained language model can and cannot do
on a given task. In this section, we show that model training itself is governed by the same primitives. The eigenspectrum of the dependency kernel, averaged over the task mixture in the training data, influences model training in two ways. Firstly, during the pretraining phase it determines the exponents of a resource-scaling curve that expresses empirical loss as a function of the amount of training data, model parameter count, and the conditional entropy $H(Y|X)$. Secondly, when a trained model is finetuned on a specific task, it helps quantify loss of performance on other tasks the model was originally trained for. 

\subsection{Scaling law for model training}
\label{sec:scaling}

We derive a power-law relationship (Theorem~\ref{thm:decomp}) between cross-entropy loss $\mathcal L$ and the number of model parameters $N$ and the number of training tokens $D$.  The empirically validated version is known as the Chinchilla scaling law~\citep{kaplan2020, hoffmann2022}:
\begin{align}
    \mathcal L(N,D) &= \frac{A}{N^\alpha} + \frac{B}{D^\beta} + E,
    \label{eq:chinchilla}
\end{align}
where $E$ is an irreducible error term, and $\alpha$, $\beta$, $A$, $B$ are fitted constants. Our scaling law is a slight generalization of this form, and contains the same three summands. Eq.~\eqref{eq:chinchilla} arises as a special case. We start with the machinery below, then use them to identify $\beta, \alpha$, and $E$ in turn.

\subsubsection{Task mixture eigenspectrum}
Since an LLM is trained on a distribution $\calT$ over tasks, its scaling behavior depends not on any single task's dependency kernel, but on the shared structure across tasks.  

\begin{definition}
\label{def:task_cov}
The \emph{task mixture kernel} is the task-averaged dependency kernel $\bar{K}(t,t') = \E_{\tau \sim \calT}[K_Y^\tau(t,t')]$, with eigenvalues $\{\bar{\sigma}_k\}_{k \geq 1}$ in non-increasing order.
\end{definition}

The eigenspectrum of $\bar{K}$ captures which dependency patterns are shared across tasks (large $\bar{\sigma}_k$) and which are specific to individual task families (small $\bar{\sigma}_k$). We make two assumptions on it.

\begin{assumption}[Power-law mixture spectrum]
\label{asmp:power}
The eigenvalues of $\bar K$ are nonnegative, with $\bar{\sigma}_k \sim \bar{\sigma}_1 k^{-\nu_{\calT}}$ for a spectral decay exponent $\nu_{\calT} > 0$.
\end{assumption}

\begin{assumption}[Average target regularity]
\label{asmp:target}
$\E_\tau[\bar{f}_k^{\tau\,2}] \sim \bar{f}_0^2 k^{-\bar{\mu}}$ for a regularity exponent $\bar{\mu} > 1$, where $\bar{f}_k^\tau = \langle f^\tau, \bar{e}_k \rangle$ is the projection of task $\tau$'s prediction target onto the $k$-th eigenmode of $\bar{K}$.
\end{assumption}

The power-law assumptions are empirically well-motivated. Eigenspectra of data covariance operators consistently follow power laws across domains, with exponents varying 
by dataset~\citep{canatar2021}. 
Prediction targets that are smooth relative to the data covariance (i.e., well-approximated by their leading spectral components) are a standard condition in nonparametric regression, and power-law decay of projections is widely observed in practice~\citep{bordelon2020}.

\subsubsection{The data exponent ($\beta$)}
At each position $t$, the autoregressive model must learn the prediction function $f_t(y_{<t}, X) = P(y_t \mid y_{<t}, X)$.  Viewed as an element of a reproducing kernel Hilbert space, this function's complexity relative to the training distribution is captured by a data covariance operator $\bar{\Sigma}_t = \E_{(\tau, y_{<t}, X) \sim P} [\phi(y_{<t}, X) \otimes \phi(y_{<t}, X) ]$, where the expectation is over the training distribution, which draws task $\tau$ from $\calT$ and then draws $(y_{<t}, X)$ from $P^\tau$.

\begin{theorem}
\label{thm:beta}
Suppose that $\bar\Sigma_t$ shares the eigenbasis of $\bar K$ with
eigenvalues comparable to those of $\bar K$ uniformly across positions
$t$, 
and that the per-mode estimation error follows that of a kernel ridge regression. Then, given Assumptions~\ref{asmp:power} and~\ref{asmp:target}, the estimation component of the cross-entropy loss satisfies $\mathcal L_{\mathrm{est}}(D) = B/D^{\,\beta} + o(D^{-\beta})$ with
\begin{equation*}
    \beta \;=\; \frac{\bar{\mu} - 1}{\nu_{\calT}}.
\end{equation*}
\end{theorem}

The numerator $\bar{\mu} - 1$ measures how quickly the average prediction signal decays across eigenmodes; the denominator $\nu_{\calT}$ measures how quickly the mixture data variance decays.  When the signal is concentrated and the spectrum is steep, convergence is fast (large $\beta$); when the signal is diffuse and the spectrum is flat, convergence is slow (small $\beta$).

\subsubsection{The capacity exponent $(\alpha)$}
The data exponent $\beta$ governs how quickly additional data reduces error.  The capacity exponent $\alpha$ governs how quickly additional model capacity reduces error. Both operate on the same task-mixture spectrum (Assumptions~\ref{asmp:power}--\ref{asmp:target}). To prove a similar result for the capacity exponent, the only additional ingredient is the relationship between parameters and representable modes.

\begin{theorem}
\label{thm:alpha}
Suppose a model with $N$ parameters has a representational budget of $\kappa N^{1/d}$ units, and representing eigenmode $k$ of $\bar{K}$ costs $\bar{\sigma}_k^{-1}$ units. Then, given Assumptions~\ref{asmp:power} and~\ref{asmp:target}, the approximation component of the cross-entropy loss satisfies $\mathcal L_{\mathrm{approx}}(N) = A/N^{\,\alpha} + o(N^{-\alpha})$ with
\begin{equation*}
    \alpha \;=\; \frac{\bar{\mu}-1}{d\,(\nu_{\calT}+1)}.
\end{equation*}
\end{theorem}

The exponents in Theorems \ref{thm:beta} and \ref{thm:alpha} are related as $\alpha/\beta = \nu_{\calT}/(d(\nu_{\calT}+1))$. Both are governed by the same spectral quantities $(\bar{\mu}, \nu_{\calT})$; they differ through the architectural factor $d$ and a spectral correction $\nu_{\calT}/(\nu_{\calT}+1)$.  This asymmetry arises because each eigenmode is learned independently from data but eigenmodes share a finite parameter budget, so representing one mode leaves fewer parameters for others.  For large $\nu_{\calT}$, the correction is negligible and $\alpha \approx \beta/d$.  The corrected Chinchilla estimates of~\citet{besiroglu2024} give $\alpha \approx 0.348$ and $\beta \approx 0.366$, yielding $\alpha/\beta \approx 0.95$, consistent with $d \approx 1$ and moderately large $\nu_{\calT}$.

\subsubsection{The irreducible term and complete decomposition}
The cross-entropy loss of any model $Q_\theta \equiv Q_{Y|X}$ admits the exact identity
\begin{equation}
    \mathcal{L}(N,D) \;=\; H(Y \mid X) \;+\; \E_X\bigl[\KL(P_{Y|X} \,\|\, Q_{Y|X})\bigr],
    \label{eq:loss_decomp}
\end{equation}
where $\mathcal{L}$ denotes the expected cross-entropy loss to distinguish it from the sequence length.  The first term depends only on the task distribution, and the second is the total reducible error. The Bayes-optimal predictor $Q^* = P_{Y|X}$ achieves $\mathrm{KL}(P_{Y|X} \| Q_{Y|X}) = 0$ pointwise. Thus, the irreducible floor $H(Y|X)$ is precisely the Bayes-optimal loss. The following result confirms that this floor is tight.

\begin{proposition}
\label{prop:irred_id}
For all $N, D$, $\mathcal{L}(N,D) \geq H(Y|X)$, with $\mathcal{L}(N,D) \to H(Y|X)$ as $N, D \to \infty$.
\end{proposition}

Theorems~\ref{thm:beta} and~\ref{thm:alpha} characterize the rate at which the reducible error vanishes in $\mathcal{L}(N,D)$, while the irreducible error ($E$ in Eq.~\eqref{eq:chinchilla}) is identified by $H(Y|X)$. Our scaling law follows immediately.

\begin{theorem}
\label{thm:decomp}
Under Assumptions~\ref{asmp:power} and~\ref{asmp:target}, together with the setup of Theorems~\ref{thm:beta} and~\ref{thm:alpha}, we have
\begin{align}
    \mathcal{L}(N,D)
    &= \max \left( \frac{A}{N^{(\bar \mu-1)/(d(\nu_\calT + 1))}}
    \; \frac{B}{D^{(\bar \mu-1)/\nu_\calT}} \right) \notag\\
    &  \quad + H(Y \mid X) + o \left(\max(N^{-\alpha}, D^{-\beta})\right).
    \label{eq:loss_max_main}
\end{align}
\end{theorem}

Performance is limited by the resource that is more scarce. When $A N^{-\alpha} \gg B D^{-\beta}$, capacity is the bottleneck and additional data yields negligible returns, while when $B D^{-\beta} \gg A N^{-\alpha}$, data is the bottleneck and additional parameters yield negligible returns. The empirically-fitted Chinchilla law follows as a special case along the compute-optimal frontier.

\begin{corollary}
\label{cor:additive}
When $N$ and $D$ are scaled so that $A\,N^{-\alpha} \asymp B\,D^{-\beta}$, the loss satisfies
\begin{align}
    \mathcal{L}(N,D) &= \frac{A}{N^{\alpha}} + \frac{B}{D^{\beta}} + H(Y|X) \notag\\
    & \quad + o\bigl(\max(N^{-\alpha}, D^{-\beta})\bigr).
    \label{eq:decomp}
\end{align}
\end{corollary}

The $\max$ form~\eqref{eq:loss_max_main} makes a prediction beyond the Chinchilla law: there is no benefit to scaling one resource far beyond the other. This justifies the compute-optimal strategy of~\citet{hoffmann2022}, which balances $N$ and $D$. It allocates just enough parameters to represent the modes that available data can learn, and just enough data to learn the modes that the available parameters can represent.


The irreducible term $H(Y|X)$ connects the scaling laws above back to $R^*$. With task identity for input-output pairs given, it decomposes as
\begin{align}
H(Y|X) &= \mathbb{E}_{\tau\sim\mathcal{T}}[H_\tau(Y \mid X)] \notag\\
&= \mathbb{E}_{\tau\sim\mathcal{T}} (1-R^*_\tau) H_\tau(Y) \notag\\
&= \underbrace{\mathbb{E}_{\tau\sim\mathcal{T}}\!\big[\delta_{r,\tau} H_\tau(Y)\big]}_{\text{resolvable floor}}
+ \underbrace{\mathbb{E}_{\tau\sim\mathcal{T}}\!\big[\delta_{u,\tau} H_\tau(Y)\big]}_{\text{subjective floor}}.
\label{eq:task-decomp}
\end{align}
This separates the three quantities a firm needs to consider for training and deploying an LLM across tasks. Scaling $N$ and $D$ reduces $\max(AN^{-\alpha}, BD^{-\beta})$ and leaves both floors unchanged (Theorem~\ref{thm:decomp}). Auxiliary context reduces the resolvable floor and leaves the subjective floor unchanged (Theorem~\ref{thm:context}). Consequently, if a task $\tau$ requires a cross-entropy loss below $\delta_{u,\tau}H_\tau(Y)$ to meet the firm's quality requirement, no combination of scaling and context enrichment achieves it, and the appropriate response is to redefine the task so that the subjective component is removed from the automated portion.

\subsection{Fine-tuning as spectral reallocation}
The scaling law of Theorem~\ref{thm:decomp} governs a generalist model trained on a task mixture $\calT$.  A natural question is what happens when such a model is finetuned to a single task, 
as a firm would do when adapting a generic LLM for a business-relevant application. It
is a robust empirical finding that this improves performance on the
target task while degrading it on tasks the model previously handled
well, a phenomenon known as \emph{catastrophic forgetting}
\citep{mccloskey1989, kirkpatrick2017}. We posit that finetuning redistributes a fixed representational budget across the eigenmodes of the dependency kernel, and a task loses performance in proportion to how much of its own structure lived in the modes that were given up. To formalize this, we introduce a measure of structural similarity between tasks.

\begin{definition}
\label{def:task_align}
The \emph{alignment} between tasks $\tau_1$ and $\tau_2$ is $\rho(\tau_1,\tau_2) = \mathrm{tr}(K_Y^{\tau_1} K_Y^{\tau_2}) / (\|K_Y^{\tau_1}\|_F \|K_Y^{\tau_2}\|_F) \in [0,1]$.  High alignment means two tasks share dependency structure; low alignment means their kernels are approximately orthogonal.
\end{definition}

During pretraining, the model serves the task distribution $\calT$, and Theorem~\ref{thm:alpha} gives the capacity exponent $\alpha = (\bar{\mu}-1)/(d(\nu_{\calT}+1))$.  The model's $M(N)$ representable modes are distributed across the eigenmodes of the mixture kernel $\bar{K} = \E_{\tau \sim \calT}[K_Y^\tau]$, with capacity allocated in proportion to the eigenvalues $\{\bar{\sigma}_k\}$. Fine-tuning on $\tau^*$ reoptimizes the model's parameters under the loss for $\tau^*$ alone. 

\begin{proposition}
\label{prop:finetune}
Fine-tuning on task $\tau^*$ tilts the effective eigenspectrum from $\bar{K}$ toward $K_Y^{\tau^*}$, shifting the spectral parameters from $(\nu_{\calT}, \bar{\mu})$ toward $(\nu_{\tau^*}, \mu_{\tau^*})$
.  The performance loss on task $\tau \neq \tau^*$ scales with $\sqrt{1 - \rho(\tau, \tau^*)^2}$.
\end{proposition}

Fine-tuning on a single task $\tau^*$ tilts the dependency kernel eigenspectrum: modes aligned with $\tau^*$ receive greater weight while modes serving unrelated tasks are suppressed. The effective spectral parameters shift from the mixture values $(\nu_{\calT}, \bar{\mu})$ toward the task-specific $(\nu_{\tau^*}, \mu_{\tau^*})$. Catastrophic forgetting is the necessary cost of this. The task families served by the suppressed modes lose their representational support when those modes are reallocated. According to Proposition~\ref{prop:finetune}, the severity of forgetting for a task $\tau \neq \tau^*$ scales with the misalignment between the dependency kernels of $\tau$ and $\tau^*$, so that tasks with similar structure share modes and are partially preserved, while orthogonal tasks degrade substantially.

\begin{remark}
In light of Eq.~\eqref{eq:task-decomp}, the fine-tuning tradeoff above is different across tasks. Fine-tuning on $\tau^*$ does not change $H_{\tau^*}(Y\mid X)$; it can only reduce the reducible error on $\tau^*$. The gain from fine-tuning is therefore capped by this quantity, while the loss on a retained task $\tau$ scales as $\sqrt{1-\rho(\tau,\tau^*)^2}$ irrespective of it. For an unverifiable task $\tau^*$ on which the pretrained model is already close to a high floor $\delta_{u,\tau^*}H_{\tau^*}(Y)$, the attainable gain is small but the potential loss is large. Fine-tuning is justified when the pretrained model's reducible gap on $\tau^*$ exceeds the forgetting cost summed over retained tasks, weighted by their value to the firm. This condition fails for highly ambiguous tasks, so the firm should neither scale nor specialize toward them.
\end{remark}

\section{Conclusion}
\label{sec:discussion}

We have shown that the performance of a generative language model on a task is bounded by two properties of the task itself. The reliability ceiling $R^*$ bounds what any model can achieve from the observable input, and the dependency kernel $K_Y$ governs how far autoregressive
generation falls below that ceiling as the output lengthens. The scaling law of Theorem~\ref{thm:decomp} follows from the same two primitives applied to the task mixture, so that the exponents practitioners fit empirically have a structural origin in the spectral decay of shared dependency structure and the concentration of prediction signal in its leading modes.

The framework explains the ``jagged frontier'' of AI capability, the observation that AI assistance improves performance on some tasks while degrading it on others \citep{dellacqua2023}. The frontier follows each task's reliability ceiling and dependency structure, and superficially similar tasks can fall on opposite sides of it (``write a legal contract for $X$'' versus ``write a persuasive legal argument''). In the related debate on whether LLMs are ``hitting a wall,'' both
sides are partially correct. Scaling works, but at rates that vary across task mixtures and toward ceilings that range from near-perfect (arithmetic) to well below (creative writing). The max form of the scaling law (Eq.~\ref{eq:decomp}) adds a further prediction: scaling one resource far beyond the other yields negligible returns, so scaling maximalism without architecture-level changes will plateau at the level set by the binding resource.

Returning to the questions posed in Section~\ref{sec:setup}, our results carry three implications for a firm deploying LLMs. First, the task taxonomy sorts candidate tasks into three treatments. Fully verifiable tasks can be automated directly. Ideally verifiable tasks become automatable once
the firm supplies the context that closes $\delta_r$, through retrieval over internal documents, tool integration, or tighter specification; Theorem~\ref{thm:context} limits the return on this investment by $\delta_r$, so it can be sized before it is made. Unverifiable tasks retain a floor $\delta_u$ that no model choice or context engineering removes, so the automated portion of such a task should be scoped so that the subjective component stays with a human evaluator. Second, Propositions~\ref{prop:dep} and \ref{prop:verify} show that the effectiveness of verification depends on whether correctness is checkable from a prefix, which is a property of the output format as much as of the task. Constraining outputs to structured formats, such as schemas, typed fields, or templates, moves a task toward a banded kernel, and the results predict that this both
slows degradation and makes intermittent verification effective. Whether decomposing a task into subtasks has the same effect is not established here. Third, Proposition~\ref{prop:finetune} gives a criterion for the choice between fine-tuning one shared LLM vs. maintaining separate LLMs for separate task sets. The alignment parameter $\rho_{\tau,\tau^*}$ is estimable from output data without access to the model internals, so this criterion can be evaluated before fine-tuning is undertaken.

The results also bear on performance evaluation of LLMs. Proposition~\ref{prop:misspec} shows that model rankings on tasks with $R^*<1$ are unstable at practical benchmark sizes, so aggregate leaderboard scores that mix tasks with different ceilings are unreliable guides to model selection. We recommend that evaluation on such tasks use entropy-aware metrics with multiple acceptable references per instance, and that firms evaluate candidate models on business-specific benchmark datasets rather than public leaderboards.

\paragraph{Limitations and future work.}
In our proposal, the reliability ceiling and dependency kernel are not directly observable. While the numerical illustrations use reasonable proxies for each, the finite-sample estimation theory for these quantities is an open problem. 
The deployment implications above are qualitative; an economic model that prices context addition mechanisms, verification mechanisms, and task forgetting against the payoff from task completion, and chooses among them subject to a budget, would turn them into an optimization problem. Further directions include departures from the power-law regime \citep{caballero2023brokenneuralscalinglaws}, scaling laws for in-context learning \citep{arora2025bayesianscalinglawsincontext} and compression \citep{panferov2026unified}, the evolution of the dependency kernel under non-stationary training, and the extension of the framework to non-autoregressive architectures.


\section*{Acknowledgements}
This research is supported by the Indian Institute of Management Bangalore Young Faculty Research Grant. The author thanks Tirthatanmoy Das, Anand Deo, Domenic Rosati, Kush Varshney, Lav Varshney, and Dootika Vats for their thoughts and feedback on early drafts. 

\bibliographystyle{abbrvnat}
\bibliography{reference}

\appendix

\section{Task Verifiability and Computational Complexity}
\label{app:complexity}

The task taxonomy of Section~\ref{sec:formalism} admits a complexity-theoretic interpretation, connecting fully verifiable tasks to $\mathrm{P}$, ideally verifiable tasks to $\mathrm{NP}$ (where $C_r$ serves as a certificate), and unverifiable tasks to problems without efficient verifiers.  We make this precise below, noting an important caveat: the mapping is between our information-theoretic categories and the \emph{structural properties} of complexity classes, not a formal equivalence.  In particular, a fully verifiable task ($H(Y|X) = 0$) has a deterministic input-output mapping, but this mapping may itself be computationally intractable
.

\paragraph{Fully verifiable tasks and $\mathrm{P}$.}  When $\delta_r = \delta_u = 0$, the output $Y = f(X)$ is a deterministic function of the input.  A verification oracle $V(x,y) = \mathbb{1}[y = f(x)]$ exists trivially, and if $f$ is polynomial-time computable, the task is in $\mathrm{P}$ in the standard sense: both generation and verification are efficient.

\paragraph{Ideally verifiable tasks and $\mathrm{NP}$.}  When $\delta_u = 0$ and $\delta_r > 0$, the output is deterministic given $(X, C_r)$ but not given $X$ alone.  The resolvable context $C_r$ plays the role of an $\mathrm{NP}$ certificate: a verifier $V(x, y, c_r) = \mathbb{1}[y = g(x, c_r)]$ can check the correctness efficiently given the certificate, even when finding $y$ (or $c_r$) from $x$ alone is hard. Reinforcement learning with verifiable rewards (RLVR) exploits this structure: the verifier provides a clean reward signal because $\delta_u = 0$ guarantees no noise from subjective disagreement.

\begin{proposition}
\label{prop:rlvr_clean}
If $\delta_u = 0$ and a polynomial-time verifier $V(x,y)$ exists, then the reward signal $r(y) = V(x,y)$ has zero label noise from subjective disagreement.
\end{proposition}

\paragraph{Proof.}
Since $\delta_u = 0$, the set of correct outputs $\{y : V(x,y) = 1\}$ is well-defined for each $x$ and does not depend on the evaluator.  The reward $r(y) = V(x,y)$ is therefore a deterministic function of $(x, y)$.  In contrast, when $\delta_u > 0$, any reward derived from human preferences inherits the irreducible disagreement encoded in $C_u$: the reward for the same $(x, y)$ pair varies across annotators, introducing label noise with variance proportional to $\delta_u$.
\hfill\ensuremath{\square}

The pass@$k$ metric further reveals this $\mathrm{NP}$-like structure.  For a model with per-sample success probability $p$, pass@$k = 1 - (1-p)^k$.  Under oracle verification, coverage grows as $1 - e^{-pk}$ and can approach $1$ with enough samples.  Under majority voting, coverage stagnates because the model cannot distinguish correct from plausible-but-incorrect outputs.  The growing gap between oracle coverage and majority-vote accuracy, documented by~\citet{brown2024monkeys} across MATH, SWE-bench, and CodeContests, is the empirical signature of this structure: solutions exist in the model's output distribution but can only be identified with a verifier.

\paragraph{Unverifiable tasks and the absence of efficient verifiers.}  When $\delta_u > 0$, no deterministic verifier can achieve perfect accuracy on the task distribution.

\begin{proposition}
\label{prop:no_verifier}
If $\delta_u > 0$, then no deterministic verifier $V: \calX \times \calV^L \to \{0,1\}$ can achieve perfect accuracy on the task distribution.
\end{proposition}

\paragraph{Proof.}
Fix an evaluator's realization $C_u = c_u$ and let $\mathcal A(x, c_r,
c_u) := \{ y : y \text{ is correct for } x \text{ under } c_u \}$ be the
set of outputs that evaluator accepts, so that correctness of $y$ is
the event $y \in \mathcal A(X, C_r, C_u)$. A deterministic verifier
$V : \mathcal X \times \calV^L \to \{0,1\}$ observes only $(x, y)$, so
its accepted set $\mathcal A_V(x) := \{ y : V(x,y) = 1 \}$ is fixed
once $x$ is given; perfect accuracy means $\mathcal A_V(X) = \mathcal
A(X, C_r, C_u)$ almost surely.

Suppose such a $V$ exists. Then membership in $\mathcal A(X, C_r,
C_u)$ is determined by $X$ alone, so $Y$ is conditionally independent
of $C_u$ given $(X, C_r)$ on the event $\{Y \in \mathcal A\}$, which
has probability one under the data-generating process: the output
drawn for a given input is correct for the evaluator who produced it.
Hence $I(C_u; Y \mid X, C_r) = 0$, contradicting $\delta_u > 0$.

Equivalently, $\delta_u > 0$ means $\mathcal A(x, c_r, \cdot)$ varies
with $c_u$ on a set of positive measure. On that set, any fixed
$\mathcal A_V(x)$ disagrees with $\mathcal A(x, c_r, c_u)$ for at least
one realization of $c_u$, so $V$ misclassifies some
$(x, y)$ pairs with positive probability. Note that the failure is not that
multiple outputs are acceptable, which a verifier can accommodate, but
that which outputs are acceptable is not a function of the observable
input.
\hfill\ensuremath{\square}



\section{Proofs of Results}
\label{app:proofs}

\subsection{Proof of Theorem~\ref{thm:ceiling}}

The proof applies the data processing inequality (DPI) to the Markov chain established in~\eqref{eq:markov}.

The Markov chain $Y \leftrightarrow (X,C) \leftrightarrow X \leftrightarrow T \leftrightarrow \hat{Y}$ yields three applications of the DPI:

\emph{Step 1: Decoder loss.}  Since $\hat{Y}$ is a stochastic function of $T$ (i.e., $\hat{Y} \perp (X, Y, C) \mid T$), and $\phi(\hat{Y})$ is a deterministic function of $\hat{Y}$, the DPI gives
$$
    I(\phi(\hat{Y}); Y) \;\leq\; I(\hat{Y}; Y) \;\leq\; I(T; Y).
$$
The first inequality holds because $\phi(\hat{Y})$ is a (possibly lossy) processing of $\hat{Y}$; the second because $\hat{Y}$ is generated from $T$ alone.

\emph{Step 2: Encoder loss.}  Since $T$ is computed from $X$ alone (i.e., $T \perp C \mid X$, and hence $T$ is a function of $X$ and possibly independent randomness), the DPI gives
$$
    I(T; Y) \;\leq\; I(X; Y).
$$

\emph{Step 3: Marginalization loss.}  Since $X$ is a marginal of $(X, C)$, the DPI gives
$$
    I(X; Y) \;\leq\; I((X,C); Y).
$$

Chaining Steps 1--3 yields~\eqref{eq:ceiling_chain}: $I(\phi(\hat{Y}); Y) \leq I(T; Y) \leq I(X; Y)$.

For the ceiling bound, take $\phi$ to be the identity.  Then $I(\hat{Y}; Y) \leq I(X; Y)$ for any model $Q_{Y|X}$.  Dividing both sides by $H(Y) > 0$ (which holds for any non-degenerate task) and taking the supremum over all models:
$$
    \sup_{Q_{Y|X}} \frac{I(\hat{Y}; Y)}{H(Y)} \;\leq\; \frac{I(X; Y)}{H(Y)} \;=\; \Rstar. 
$$

\subsection{Proof of Theorem~\ref{thm:context}}
\label{sec:proof_thm_context}


\textbf{Part (i).}
By the chain rule of mutual information, $I(X, \tilde C; Y) = I(X; Y) + I(\tilde C; Y \mid X)$.
Since $I(\tilde C; Y \mid X) \geq 0$ by non-negativity, $I(X, \tilde C; Y) \geq I(X; Y)$, and dividing by $H(Y)$ gives $\tilde{R}^* \geq \Rstar$.  Equality holds if and only if $I(\tilde C; Y \mid X) = 0$, i.e., $\tilde C \perp Y \mid X$.  This occurs when $\tilde C$ is a deterministic function of $X$ or otherwise carries no information about $Y$ beyond what $X$ already provides.

\textbf{Part (ii).}
We show $I(\tilde C; Y \mid X) \leq I(C_r; Y \mid X) = \delta_r \cdot
H(Y)$.  Recall the decomposition $C = (C_r, C_u)$ with $Y = g(X, C_r,
C_u)$.  Given $(X, C_r)$, the output $Y$ is a function of $C_u$, so the
assumption $\tilde C \perp C_u \mid (X, C_r)$ implies $\tilde C \perp Y
\mid (X, C_r)$, i.e.\ $I(\tilde C; Y \mid X, C_r) = 0$. By the chain
rule of mutual information,
$$
    I(\tilde C; Y \mid X) \;\leq\; I(\tilde C, C_r; Y \mid X)
    \;=\; I(C_r; Y \mid X) + I(\tilde C; Y \mid X, C_r)
    \;=\; I(C_r; Y \mid X) \;=\; \delta_r \cdot H(Y),
$$
where the first inequality is monotonicity of mutual information in
its first argument.  Dividing by $H(Y)$:
$$
    \tilde{R}^* - \Rstar = \frac{I(\tilde C; Y \mid X)}{H(Y)} \leq \delta_r.
$$

\textbf{Part (iii).}
Let $\tilde C_1, \ldots, \tilde C_n$ be a sequence of auxiliary variables.  Define $\tilde{R}^*_k = I(X, \tilde C_1, \ldots, \tilde C_k; Y)/H(Y)$.  By the chain rule,
$$
    \tilde{R}^*_{k+1} - \tilde{R}^*_k = \frac{I(\tilde C_{k+1}; Y \mid X, \tilde C_1, \ldots, \tilde C_k)}{H(Y)} \geq 0,
$$
so the sequence $\{\tilde{R}^*_k\}$ is nondecreasing.  It is bounded above by $I(X, C_r; Y)/H(Y) = 1 - \delta_u$ from part~(ii), since the cumulative information from all auxiliary variables cannot exceed the total resolvable information $I(C_r; Y \mid X)$.  When $\tilde C$ collectively resolves all of $C_r$---i.e., $C_r$ is a measurable function of $(X, \tilde C_1, \ldots, \tilde C_n)$---then $I(X, \tilde C; Y) = I(X, C_r; Y)$ and
$$
    \tilde{R}^*_{\max} = \frac{I(X, C_r; Y)}{H(Y)} = 1 - \delta_u.
$$
This is exactly $1$ when $\delta_u = 0$ (ideally verifiable tasks) and strictly less than $1$ when $\delta_u > 0$ (unverifiable tasks).

\subsection{Proof of Proposition~\ref{prop:misspec}}
\label{sec:proof_prop_misspec}


Write $\bar s_j = \mu_j + \epsilon_j$ and $\bar s'_j = \mu_j +
\epsilon'_j$ for the two runs, with $\epsilon_j, \epsilon'_j$
independent, zero-mean, of common variance $\sigma^2_\epsilon$, and
$\sigma^2_\mu > 0$ the variance of $\mu_j$ across models. Then
$\mathrm{Corr}(\bar s, \bar s') = \sigma^2_\mu/(\sigma^2_\mu +
\sigma^2_\epsilon)$, which is strictly less than one iff
$\sigma^2_\epsilon > 0$. Since $R^* < 1$, $H(Y \mid X) > 0$, so
$H(Y \mid X = x) > 0$ on a set of inputs of positive measure, and on
that set $\max_y P(y \mid x) < 1$. For such $x$, the match probability
$p = \sum_y Q_j(y \mid x) P(y \mid x)$ satisfies $0 < p \le \max_y
P(y \mid x) < 1$, so the exact-match score is a non-degenerate
Bernoulli and has positive conditional variance. Hence
$\sigma^2_\epsilon = M^{-2}\sum_m \mathrm V(s_j^{(m)}) > 0$.

\subsection{Proof of Proposition~\ref{prop:fidelity-bridge}}
\label{sec:proof_prop_fidelity_bridge}

Write $F := F_\theta$ and let $E := \mathbf 1[\, d( \hat Y, Y ) > \eta
\,]$, so $\Prob( E = 0 ) = F$. All variables are discrete. Expanding
$H( Y, E \mid \hat Y )$ two ways and dropping a nonnegative term,
$$
H( Y \mid \hat Y ) \;\le\; H( E ) \;+\; H( Y \mid \hat Y, E ) \;\le\;
\log 2 \;+\; F\, H( Y \mid \hat Y, E = 0 ) \;+\; ( 1 - F )\, H( Y \mid
\hat Y, E = 1 ).
$$
Conditionally on $\hat Y = \hat y$ and $E = 0$, the variable $Y$ is
supported on the tolerance ball $\{ y : d( \hat y, y ) \le \eta \}$, so
$H( Y \mid \hat Y, E = 0 ) \le \log N_\eta$; and $H( Y \mid \hat Y, E =
1 ) \le \log |\mathcal V|^L = L \log |\mathcal V|$. Hence, using $F \le
1$ on the middle term,
$$
H( Y \mid \hat Y ) \;\le\; \log 2 \;+\; \log N_\eta \;+\; ( 1 - F )\, L
\log |\mathcal V| .
$$
For the lower bound, the generator's sampling randomness is independent
of $Y$ given $X$, so $Y - X - \hat Y$ is a Markov chain and the data
processing inequality gives $I( \hat Y; Y ) \le I( X; Y ) = \Rstar\, H(
Y )$. Therefore
$$
H( Y \mid \hat Y ) \;=\; H( Y ) - I( \hat Y; Y ) \;\ge\; H( Y ) \big( 1
- \Rstar \big).
$$
Combining the two displays,
$$
( 1 - F )\, L \log |\mathcal V| \;\ge\; H( Y ) - \Rstar\, H( Y ) -
\log( 2 N_\eta ) \;\ge\; H( Y ) - \log( 2 N_\eta ) - \Rstar\, L \log
|\mathcal V| ,
$$
where the last step uses $H( Y ) \le L \log |\mathcal V|$ and $\Rstar
\ge 0$. 

\subsection{Proof of Proposition~\ref{prop:amplification}}
\label{sec:proof_prop_amplification}

Let $E_t := \{ z : d( [\hat y_{<t}\, z], y_{\le t} ) > \eta \}$, so that
$\epsilon_t = \Prob_z[ E_t \mid \hat y_{<t} ]$, and split the tokens by
what they cost at the reference prefix:
$$
A := \big\{ z : d( [y_{<t}\, z], y_{\le t} ) > \eta \big\}, \qquad
B := \big\{ z : \eta - r_t < d( [y_{<t}\, z], y_{\le t} ) \le \eta
\big\} .
$$
These are disjoint, with $\Prob_z[ A \mid \hat y_{<t} ] = \epsilon^*_t$
and $\Prob_z[ B \mid \hat y_{<t} ] = g_t$, and
$$
A \cup B \;=\; \big\{ z : d( [y_{<t}\, z], y_{\le t} ) > \eta - r_t
\big\} .
$$
For $z \in A \cup B$, Assumption~\ref{ass:budget} gives
$$
d\big( [\hat y_{<t}\, z], y_{\le t} \big) \;\ge\; r_t + d\big(
[y_{<t}\, z], y_{\le t} \big) \;>\; r_t + ( \eta - r_t ) \;=\; \eta ,
$$
so $A \cup B \subseteq E_t$ and $\epsilon_t \ge \epsilon^*_t + g_t$.
The inclusion holds token by token and uses no property of any
distribution; both probabilities are taken under the law that actually
emits $\hat y_t$, so no comparison between two distributions occurs.

Note that the above covers the case $r_t > \eta$. There $\eta - r_t < 0$, so $B$ contains every token with $d([y_{<t}\, z], y_{\le t} ) \le \eta$ and $g_t = 1 - \epsilon^*_t$. The bound then reads $\epsilon_t \ge 1$, which recovers the fact that once the drift exceeds the tolerance no token can bring the output back.


Let $\mathcal F_t := \sigma( X, Y, \hat y_{\le t} )$ and
$$
A_t := \big\{ d( \hat y_{\le t}, y_{\le t} ) \le \eta \big\}, \qquad
B_t := \bigcap_{s \le t} A_s, \qquad B_0 := \Omega .
$$
 
\emph{Step 0 (drift never decreases).} Assumption~\ref{ass:distance},
applied with $u = \hat y_{\le s}$ and $z = \hat y_{s+1}$, gives $d(
\hat y_{\le s+1}, y_{\le s+1} ) \ge d( \hat y_{\le s}, y_{\le s} )$, so
$s \mapsto d( \hat y_{\le s}, y_{\le s} )$ is nondecreasing. Hence $A_L
\subseteq A_s$ for every $s \le L$, so $\{ d( \hat Y, Y ) \le \eta \} =
A_L = B_L$, and $r_t \le \eta$ holds on $B_{t-1}$, so part (i) and the
floor apply at every step at which the output is still within
tolerance.
 
\emph{Step 1 (one step).} On $B_{t-1}$, part (i) and the floor give
\begin{align}
\mathbb P\big( A_t \,\big|\, \mathcal F_{t-1} \big) 
&\;=\; 1 - \epsilon_t \notag\\
&\;\le\; 1 - \epsilon^*_t - g_t \notag\\
&\;\le\; 1 - \epsilon - g_t \notag\\
&\;\le\; e^{-( \epsilon + g_t )},
\tag{$\ast$}
\end{align}
using $1 - x \le e^{-x}$. Both $\epsilon^*_t$ and $g_t$ are $\mathcal
F_{t-1}$-measurable, being functions of $X$, $Y$, and $\hat y_{<t}$.
 
\emph{Step 2 (part (ii)).} Define
$$
M_t \;:=\; \mathbb 1_{B_t}\, \exp\Big( \epsilon\, t + \sum_{s \le t}
g_s \Big), \qquad M_0 := 1 .
$$
The exponential factor is $\mathcal F_{t-1}$-measurable and $\mathbf
1_{B_t} = \mathbb 1_{B_{t-1}} \mathbb 1_{A_t}$, so
$$
\mathbb E\big[ M_t \,\big|\, \mathcal F_{t-1} \big] \;=\;
e^{\,\epsilon t + \sum_{s \le t} g_s}\; \mathbb 1_{B_{t-1}}\, \mathbb
P\big( A_t \,\big|\, \mathcal F_{t-1} \big) \;\le\; M_{t-1}
$$
by $(\ast)$ on $B_{t-1}$; off $B_{t-1}$ both sides vanish. Hence $(
M_t )_{t \le L}$ is a nonnegative supermartingale and $\mathbb E[ M_L
] \le M_0 = 1$. On $B_L \cap \{ G_L \ge D \}$ we have $M_L \ge
e^{\epsilon L + D}$, so Markov's inequality gives
$$
\mathbb P\big[ B_L \cap \{ G_L \ge D \} \big] \;\le\; e^{-( \epsilon L
+ D )}\, \mathbb E[ M_L ] \;\le\; e^{-( \epsilon L + D )},
$$
which is part (ii) by Step 0.

\emph{Step 3 (part (iii)).} By Step 2, $\mathbb E\big[ \mathbb 1_{B_L}
e^{\epsilon L + G_L} \big] = \mathbb E[ M_L ] \le 1$. Writing $\mathbb
1_{B_L} = \mathbb 1_{B_L} e^{(\epsilon L + G_L)/2}\, e^{-(\epsilon L +
G_L)/2}$ and applying the Cauchy--Schwarz inequality,
$$
\mathbb P( B_L )
\;\le\; \Big( \mathbb E\big[ \mathbb 1_{B_L}\, e^{\epsilon L + G_L}
\big] \Big)^{1/2}
\Big( \mathbb E\big[ \mathbb 1_{B_L}\, e^{-\epsilon L - G_L} \big]
\Big)^{1/2}
\;\le\; e^{-\epsilon L/2}\, \Big( \mathbb E\big[ e^{-G_L} \big]
\Big)^{1/2}
\;=\; \gamma(L) e^{-\epsilon L/2},
$$
where the second inequality drops the indicator in the second factor.
Since $F_\theta = \mathbb P( B_L )$ by Step 0, this is part (iii).


\subsection{Proof of Proposition~\ref{prop:dep}}
\label{app:prop_dep_proof}

Say the model \emph{deviates} at step $s$ when $\hat y_s$ is not the
reference token, write $\xi_s$ for the indicator of that event, let $q_t := \Prob[\, \xi_t = 1 \mid \hat y_{<t} \,]$ be the deviation rate, $\bar q, \pi>0$ be the maximum and minimum deviation, respectively ($\bar q \geq q_t \ge \pi \,\forall\, t$), and let $N_t := \sum_{s \le t} \xi_s$. Write $a \le b$ for the implied constants in $g_t \asymp \sum_{s<t} K_Y(s,t)\, \xi_s$, that is
\begin{equation}
a \sum_{s<t} K_Y(s,t)\, \xi_s \;\;\le\;\; g_t \;\;\le\;\; b \sum_{s<t}
K_Y(s,t)\, \xi_s.
\end{equation}

Exchanging
the order of summation,
\begin{equation}
\sum_{t \le L} \sum_{s<t} K_Y(s,t)\, \xi_s \;=\; \sum_{s<L} \xi_s
\sum_{t=s+1}^{L} K_Y(s,t) ,
\label{eq:swap}
\end{equation}
so the total amplification is a sum over deviated positions, each
weighted by the kernel mass linking it to the positions that follow.

(i) For banded $K_Y$ the inner sum in \eqref{eq:swap} has at most $w$
nonzero terms, so it is at most $R := w \max_{s,t} K_Y(s,t)$, giving
$G_L \le b R\, N_L$. Since $x \mapsto e^{-x}$ is convex, Jensen's
inequality and $\mathbb E N_L = \sum_{t \le L} \mathbb E\, q_t \le
\bar q L$ give
$$
\gamma(L) \;=\; \mathbb E\, e^{-G_L} \;\ge\; \mathbb E\, e^{-b R N_L}
\;\ge\; e^{-b R\, \mathbb E N_L} \;\ge\; e^{-b R \bar q L},
$$
which is the claim with $c_1 = b R \bar q$.

(ii) Let $T := \lfloor L/2 \rfloor$, so $T \ge L/4$ and $L - T \ge L/2$
for $L \ge 2$, and let $C := \{ N_T \ge \pi T / 2 \}$.

On $C$: for every $t > T$, the lower half of the hypothesis and the
dense kernel give $g_t \ge a \kappa N_{t-1} / L \ge a \kappa N_T / L
\ge a \kappa \pi T / ( 2 L )$. Summing over the $L - T$ steps after
$T$,
$$
G_L \;\ge\; ( L - T )\, \frac{a \kappa \pi T}{2 L} \;\ge\;
\frac{L}{2}\, \cdot \frac{a \kappa \pi}{2 L} \cdot \frac{L}{4} \;=\;
\frac{a \kappa \pi}{16}\, L ,
$$
hence $\mathbb E[ e^{-G_L} \mathbb 1_C ] \le e^{-a \kappa \pi L / 16}$.

On $C^c$: since $\mathbb E[ e^{-\xi_t} \mid \mathcal F_{t-1} ] = 1 - (
1 - e^{-1} ) q_t \le \exp( -( 1 - e^{-1} ) \pi )$, iterating over $t
\le T$ gives $\mathbb E\, e^{-N_T} \le \exp( -( 1 - e^{-1} ) \pi T )$.
Markov's inequality applied to $e^{-N_T} \ge e^{-\pi T/2}$ on $C^c$
gives
$$
\Prob( C^c ) \;\le\; e^{\pi T / 2}\, \mathbb E\, e^{-N_T} \;\le\;
\exp\Big( -\pi T \big( 1 - e^{-1} - \tfrac12 \big) \Big) \;\le\;
e^{-\pi T / 8} \;\le\; e^{-\pi L / 32},
$$
using $1 - e^{-1} - \tfrac12 \ge \tfrac18$. Since $e^{-G_L} \le 1$,
$$
\gamma(L) \;=\; \left( \mathbb E\big[ e^{-G_L} \mathbb 1_C \big] + \mathbb
E\big[ e^{-G_L} \mathbb 1_{C^c} \big] \right)^2
\;\le\; \left( e^{-a \kappa \pi L / 16} + e^{-\pi L / 32} \right)^2
\;\le\; 2\, e^{-c_2 L},
$$
with $c_2 = \min\{ a \kappa \pi / 16, \; \pi / 32 \} \log 2$.

Note that the two parts of the theorem establish complementary one-sided bounds:

\medskip
\begin{center}
\begin{tabular}{lll}
    \hline
    \textbf{Kernel} & \textbf{Bound} & \textbf{Meaning} \\
    \hline
    Banded (bandwidth $w$) & $\gamma(L) \geq \exp(-c_1 L)$ & degradation at most this severe \\
    Dense (mass $\kappa$) & $\gamma(L) \leq \exp(-c_2 L)$ & degradation at least this severe \\\hline
\end{tabular}
\end{center}
\medskip

The ratio $c_2/c_1$ captures the banded-vs-dense gap, and the
regrouping \eqref{eq:swap} locates it: each deviated position is
weighted by the kernel mass linking it to the positions that follow.
In the banded case that weight is at most $R = w \max_{s,t} K_Y(s,t)$,
which is $O(1)$ relative to $L$, so $c_1 = b R \bar q$ scales linearly
in the bandwidth and the largest kernel entry. In the dense case each
weight is of order $\kappa$ however long the output is, since a
deviated position reaches all $L - s$ later positions at mass
$\kappa/L$ apiece, so $c_2$ scales linearly in $\kappa$ until it
saturates.

The difference is one of reach rather than of feedback. Under a dense
kernel a deviation raises the drift at \emph{every} subsequent
position, so the amplification $g_t$ grows with the running count of
deviations and the accumulated $G_L$ grows quadratically in the number
of early deviations, which is what forces the exponential in $L$. The
banded kernel's locality removes the reach: a deviation raises the
drift at the $w$ positions that follow it and at no others, so $G_L$
stays proportional to the number of deviations, and $\gamma(L)$ falls
by a bounded factor per token rather than compounding. The bounds above
count only this first-order effect. They do not include the further
deviations that elevated drift provokes, which compound in the dense
case and not in the banded one, so the gap they report understates the
true separation.

\subsection{Proof of Proposition~\ref{prop:verify}}
The amplification is the probability that the next token costs more
than the budget still unspent but no more than the full budget,
$g_t = \Prob_z[\, \eta - r_t < d( [y_{<t}\, z], y_{\le t} ) \le \eta
\,]$. Since $r_t^{\mathrm{ver}} \le r_t$ at every step, the interval $( \eta -
r_t^{\mathrm{ver}}, \eta ]$ is contained in $( \eta - r_t, \eta ]$, so
$g_t^{\mathrm{ver}} \le g_t$. Summing over
$t$ gives $G_L^{\mathrm{ver}} \le G_L$, and $\gamma(L) = \mathbb E\,
e^{-G_L}$ is decreasing in $G_L$, so $\gamma^{\mathrm{ver}}(L) \ge
\gamma(L)$. At $r_t^{\mathrm{ver}} = 0$ the interval is empty.
 
\subsection{Proof of Theorem~\ref{thm:beta}}

We first state and justify all assumptions required to prove this result. Assumptions \ref{asmp:power_app} and \ref{asmp:target_app} map to Assumptions \ref{asmp:power} and \ref{asmp:target} in the main paper respectively, while Assumptions \ref{asmp:align} and \ref{asmp:kernel} are mentioned briefly in the theorem statement.

\begin{assumption}[Power-law mixture spectrum]
\label{asmp:power_app}
The task covariance kernel $\bar{K}$ has nonnegative eigenvalues $\{\bar{\sigma}_k\}_{k \geq 1}$ satisfying $\bar{\sigma}_k \sim \bar{\sigma}_1 k^{-\nu_{\calT}}$ for a spectral decay exponent $\nu_{\calT} > 0$.  Large $\nu_{\calT}$ corresponds to rapid decay (locally structured tasks dominate the mixture); small $\nu_{\calT}$ close to $1$ corresponds to slow decay (the mixture spans many structurally diverse task families).
\end{assumption}

\citet{canatar2021} compute eigenspectra of neural tangent kernels on MNIST, CIFAR-10, and Fashion-MNIST, showing power-law decay with dataset-dependent exponents (their Figure~2), and verify that target projections onto these eigenbases also follow power laws (their Figure~3). Section~\ref{sec:dep_kernel} in the main paper described individual task kernels $K_Y^\tau$ in spatial terms (banded, dense, block-diagonal). The eigenspectrum provides a complementary characterization: banded kernels have rapidly decaying spectra (large $\nu_\tau$), while dense kernels have slowly decaying spectra (small $\nu_\tau$). The mixture kernel $\bar{K} = \mathbb{E}_{\tau \sim T}[K^\tau_Y]$ has eigenvalues that are a weighted 
average of the individual task kernel eigenvalues. A task family with slowly decaying eigenvalues (small $\nu_\tau$) contributes non-negligible mass at high indices $k$, preventing the mixture spectrum from decaying faster than the slowest component. Hence $\nu_T \leq \min_\tau \nu_\tau$. 

\begin{assumption}[Task--data alignment]
\label{asmp:align}
The data covariance $\bar\Sigma_t$ shares the eigenbasis $\{\bar e_k\}$
of $\bar K$, and its eigenvalues $\{\bar\lambda^{(t)}_k\}_{k \ge 1}$
are aligned with those of $\bar{K}$: there exist constants $0 < c_{\min} \leq c_{\max} < \infty$ such that $c_{\min}\bar{\sigma}_k \leq \bar{\lambda}_k^{(t)} \leq c_{\max}\bar{\sigma}_k$ uniformly across positions $t$.
\end{assumption}

This assumption holds when the training data is drawn from the same distribution that defines the task mixture---the standard in-distribution setting.  It says that patterns which are structurally important in the task mixture (large $\bar{\sigma}_k$) are also statistically common in training data (large $\bar{\lambda}_k^{(t)}$). It can fail when the training distribution is misaligned with the evaluation tasks, in which case the effective scaling exponent would be governed by the misaligned spectrum rather than the task mixture's intrinsic spectrum.

\begin{assumption}[Average target regularity]
\label{asmp:target_app}
The task-averaged projection of the prediction function onto the eigenmodes of $\bar{K}$ satisfies $\E_\tau[\bar{f}_k^{\tau\,2}] \sim \bar{f}_0^2 k^{-\bar{\mu}}$ for a regularity exponent $\bar{\mu} > 1$, where $\bar{f}_k^\tau = \langle f^\tau, \bar{e}_k \rangle$ is the projection of task $\tau$'s target onto the $k$-th eigenmode of $\bar{K}$.
\end{assumption}

The exponent $\bar{\mu}$ controls how much of the prediction signal lives in the leading modes versus the spectral tail.  When tasks are well-aligned (shared eigenbasis), $\bar{\mu}$ is close to the per-task regularity $\mu_\tau$, because each task's signal is concentrated in the same modes.  When tasks are diverse, projecting each task's target onto a shared basis spreads the signal across more modes, effectively reducing $\bar{\mu}$ relative to each task's intrinsic $\mu_\tau$.  This is the representation-level cost of multi-task learning: the shared basis is suboptimal for each individual task.

\begin{assumption}[Kernel regime]
\label{asmp:kernel}
The model's learned predictor behaves as a kernel ridge estimator on the mixture covariance: the estimation error on eigenmode $k$ is $\mathrm{err}_k(D) = \E_\tau[\bar{f}_k^{\tau\,2}] / (1 + D_{\mathrm{eff}}\bar{\lambda}_k / \sigma_{\mathrm{noise}}^2)$, where $D_{\mathrm{eff}} = D/L$ is the effective number of independent sequences.
\end{assumption}

This assumption holds exactly for kernel ridge regression and Gaussian processes, and approximately for neural networks in the lazy training regime~\citep{canatar2021, bordelon2020}.  In the heavily overparameterized interpolating regime, the kernel approximation becomes less precise. However, empirical evidence shows that the spectral decomposition of learning curves remains qualitatively correct even outside the strict kernel regime, with the power-law exponent $\beta$ preserved up to constant factors~\citep{bordelon2020}.

The cross-entropy loss of an autoregressive model averages over positions: $\mathcal \mathcal L_{\mathrm{est}}(D) = L^{-1}\sum_{t=1}^L \mathcal L_{\mathrm{est}}^{(t)}(D)$.  We prove $\mathcal L_{\mathrm{est}}^{(t)}(D) = B_t D_{\mathrm{eff}}^{-\beta} + o(D_{\mathrm{eff}}^{-\beta})$ for a generic position $t$.  The exponent $\beta$ is position-independent because Assumption~\ref{asmp:align} bounds the eigenvalues $\bar{\lambda}_k^{(t)}$ uniformly across $t$; only the prefactor $B_t$ varies with $t$.  Averaging over positions gives $\mathcal L_{\mathrm{est}}(D) = B D_{\mathrm{eff}}^{-\beta} + o(D_{\mathrm{eff}}^{-\beta})$ with $B = L^{-1}\sum_t B_t$.

\textbf{Step 1: Mode-wise error.}
Under Assumption~\ref{asmp:kernel}, the estimation error on eigenmode $k$ of $\bar{K}$ is
$$
    \mathrm{err}_k(D) \;=\; \frac{\E_\tau[\bar{f}_k^{\tau\,2}]}{1 + D_{\mathrm{eff}}\,\bar{\lambda}_k / \sigma_{\mathrm{noise}}^2}.
$$
By Assumption~\ref{asmp:align}, $\bar{\lambda}_k \geq c_{\min}\bar{\sigma}_k$, and by Assumption~\ref{asmp:target_app}, $\E_\tau[\bar{f}_k^{\tau\,2}] = \bar{f}_0^2 k^{-\bar{\mu}}(1 + o(1))$ as $k \to \infty$.  With $\bar{\sigma}_k = \bar{\sigma}_1 k^{-\nu_{\calT}}(1 + o(1))$ from Assumption~\ref{asmp:power_app}:
$$
    \mathrm{err}_k(D) \;\leq\; \frac{\bar{f}_0^2 k^{-\bar{\mu}}(1+o(1))}{1 + D_{\mathrm{eff}}\,c_{\min}\,\bar{\sigma}_1\,k^{-\nu_{\calT}}(1+o(1)) / \sigma_{\mathrm{noise}}^2}.
$$

\textbf{Step 2: Crossover index.}
Define $k^*(D)$ as the mode index at which the data term in the denominator equals $1$:
$$
    D_{\mathrm{eff}}\,c_{\min}\,\bar{\sigma}_1\,(k^*)^{-\nu_{\calT}} / \sigma_{\mathrm{noise}}^2 \;=\; 1, \qquad k^*(D) \;=\; \Bigl(\frac{D_{\mathrm{eff}}\,c_{\min}\,\bar{\sigma}_1}{\sigma_{\mathrm{noise}}^2}\Bigr)^{1/\nu_{\calT}}.
$$
Note that $k^* \to \infty$ as $D_{\mathrm{eff}} \to \infty$, so the asymptotic forms of $\bar{\sigma}_k$ and $\E_\tau[\bar{f}_k^{\tau\,2}]$ apply for all modes near and beyond the crossover.

\textbf{Step 3: Splitting the sum.}
We split $\mathcal L_{\mathrm{est}}(D) = S_{\leq}(D) + S_{>}(D)$, where
$S_{\leq}$ sums over learned modes $k \leq k^*$ and $S_{>}$ over
unlearned modes $k > k^*$.

\emph{Unlearned tail.} By Assumption~\ref{asmp:align}, write
$\bar\lambda_k = c_k \bar\sigma_k$ with $c_k \in [c_{\min}, c_{\max}]$,
and set $\theta_k := c_k / c_{\min} \in [1, \theta_{\max}]$, where
$\theta_{\max} := c_{\max}/c_{\min}$. By the definition of $k^*$ in
Step~2 and Assumption~\ref{asmp:power_app},
$$
    \frac{D_{\mathrm{eff}} \bar\lambda_k}{\sigma^2_{\mathrm{noise}}}
    \;=\; \theta_k \Big( \frac{k^*}{k} \Big)^{\nu_\calT} \big( 1 + o(1)
    \big),
$$
so for $k > k^*$ the denominator lies between $1$ and $1 +
\theta_{\max}$, and it is not asymptotically $1$: for $k$ within a
constant multiple of $k^*$ the data term is of order one. The factor
therefore stays inside the sum. With Assumption~\ref{asmp:target_app},
$$
    S_{>}(D) \;=\; \bar f_0^2 \sum_{k > k^*} \frac{k^{-\bar\mu}}{1 +
    \theta_k (k^*/k)^{\nu_\calT}} \,\big( 1 + o(1) \big),
$$
where the $o(1)$ is uniform in $k > k^*$ since $k^* \to \infty$.
Define, for $\theta > 0$,
$$
    J(\theta) \;:=\; \int_1^\infty \frac{v^{-\bar\mu}}{1 + \theta
    v^{-\nu_\calT}} \, dv ,
$$
which is finite for $\bar\mu > 1$ and satisfies $\big( (1+\theta)(\bar\mu
- 1) \big)^{-1} \le J(\theta) \le (\bar\mu - 1)^{-1}$. Since
$\theta_k \in [1, \theta_{\max}]$ and the summand is decreasing in
$\theta_k$,
$$
    \sum_{k > k^*} \frac{k^{-\bar\mu}}{1 + \theta_{\max}
    (k^*/k)^{\nu_\calT}} \;\le\; \sum_{k > k^*} \frac{k^{-\bar\mu}}{1 +
    \theta_k (k^*/k)^{\nu_\calT}} \;\le\; \sum_{k > k^*}
    \frac{k^{-\bar\mu}}{1 + (k^*/k)^{\nu_\calT}} .
$$
Each bracketing sum has a monotone summand, so by the integral
comparison with the substitution $k = k^* v$,
$$
    \sum_{k > k^*} \frac{k^{-\bar\mu}}{1 + \theta (k^*/k)^{\nu_\calT}}
    \;=\; J(\theta)\, (k^*)^{-(\bar\mu - 1)} \;+\; O\big( (k^*)^{-\bar\mu}
    \big),
$$
and the remainder is a multiplicative $O((k^*)^{-1}) = O(D_{\mathrm{eff}}
^{-1/\nu_\calT})$ correction. Hence there is $\theta^* \in [1,
\theta_{\max}]$, depending on $D$ only through lower-order terms, with
\begin{equation}
    S_{>}(D) \;=\; \bar f_0^2\, J(\theta^*)\, (k^*)^{-(\bar\mu - 1)}
    \;+\; O\big( D_{\mathrm{eff}}^{-\beta - 1/\nu_\calT} \big).
    \label{eq:tail_error}
\end{equation}
When $c_{\min} = c_{\max}$ the bracket collapses and $\theta^* = 1$.

\textbf{Step 4: Substituting the crossover index.}
Replacing $k^*$ with its expression from Step~2:
\begin{align}
    S_{>}(D) &\;=\; \frac{\bar{f}_0^2}{J(\theta^*)}\,\Bigl(\frac{D_{\mathrm{eff}}\,c_{\min}\,\bar{\sigma}_1}{\sigma_{\mathrm{noise}}^2}\Bigr)^{-(\bar{\mu}-1)/\nu_{\calT}} \notag\\
    & \quad \;+\; O\bigl(D_{\mathrm{eff}}^{-\beta - 1/\nu_{\calT}}\bigr) \nonumber \\
    &\;=\; B \cdot D_{\mathrm{eff}}^{-\beta} \;+\; O\bigl(D_{\mathrm{eff}}^{-\beta - 1/\nu_{\calT}}\bigr),
    \label{eq:tail_substituted}
\end{align}
where $\beta = (\bar{\mu}-1)/\nu_{\calT}$ and
$$
    B \;=\; \frac{\bar{f}_0^2}{\bar{\mu}-1}\,\Bigl(\frac{c_{\min}\,\bar{\sigma}_1}{\sigma_{\mathrm{noise}}^2}\Bigr)^{-(\bar{\mu}-1)/\nu_{\calT}}.
$$

Note that this identifies the position-level prefactor $B_t$.

\textbf{Step 5: Learned-mode contribution.}
For modes $k \leq k^*$, the denominator satisfies $1 \le 1 +
D_{\mathrm{eff}} \bar{\lambda}_k / \sigma^2_{\mathrm{noise}} \le 2
D_{\mathrm{eff}} \bar{\lambda}_k / \sigma^2_{\mathrm{noise}}$, so
$$
    \frac{\sigma^2_{\mathrm{noise}}}{2 D_{\mathrm{eff}} \bar\lambda_k}\,
    \mathbb{E}_\tau[\bar{f}_k^{\tau\,2}] \;\le\; \mathrm{err}_k(D)
    \;\leq\; \frac{\sigma^2_{\mathrm{noise}}}{D_{\mathrm{eff}}
    \bar\lambda_k}\, \mathbb{E}_\tau[\bar{f}_k^{\tau\,2}] .
$$
Using $c_{\min} \bar\sigma_k \le \bar\lambda_k \le c_{\max} \bar\sigma_k$
(Assumption~\ref{asmp:align}),
$$
    \frac{\bar{f}_0^2\, \sigma^2_{\mathrm{noise}}}{2 D_{\mathrm{eff}}\,
    c_{\max}\, \bar{\sigma}_1} \sum_{k=1}^{k^*} k^{\nu_{\mathcal T} -
    \bar{\mu}} \;\lesssim\; S_{\leq}(D) \;\lesssim\;
    \frac{\bar{f}_0^2\, \sigma^2_{\mathrm{noise}}}{D_{\mathrm{eff}}\,
    c_{\min}\, \bar{\sigma}_1} \sum_{k=1}^{k^*} k^{\nu_{\mathcal T} -
    \bar{\mu}},
$$
up to $(1 + o(1))$ factors from Assumptions~\ref{asmp:power_app}
and~\ref{asmp:target_app}. The two sides differ by a constant factor,
so the order of $S_{\le}(D)$ is that of $D_{\mathrm{eff}}^{-1}
\sum_{k \le k^*} k^{\nu_\calT - \bar\mu}$. We consider three cases.

\medskip
\noindent\textit{Case $\bar{\mu} > \nu_{\mathcal T} + 1$:}\;
The sum $\sum_{k=1}^{k^*} k^{\nu_{\mathcal T} - \bar{\mu}}$ converges to a constant $C_1 < \infty$ as $k^* \to \infty$, since the exponent $\nu_{\mathcal T} - \bar{\mu} < -1$.  Then $S_{\leq}(D) = \Theta(D_{\mathrm{eff}}^{-1})$.  In this regime, $\beta = (\bar{\mu}-1)/\nu_{\mathcal T} > 1$, so $D_{\mathrm{eff}}^{-\beta}$ decays \emph{faster} than $D_{\mathrm{eff}}^{-1}$: the unlearned tail $S_>$ is lower order and $S_{\leq}$ dominates.  The effective scaling exponent saturates at $\beta_{\mathrm{eff}} = 1$, regardless of how large $(\bar{\mu}-1)/\nu_{\mathcal T}$ is.  This regime (very concentrated signal with rapidly decaying spectrum) is not the empirically relevant one for language modeling, where $\beta < 1$ is consistently observed: \cite{kaplan2020} estimate $\beta \approx 0.095$, \cite{hoffmann2022} estimate $\beta \approx 0.34$, and the corrected replication of \cite{besiroglu2024} gives $\beta \approx 0.37$, all well below~1.

\medskip
\noindent\textit{Case $\bar{\mu} = \nu_{\mathcal T} + 1$:}\;
The sum is the harmonic sum $\sum_{k \le k^*} k^{-1} = \log k^* +
O(1)$, so $S_{\le}(D) = \Theta(D_{\mathrm{eff}}^{-1} \log
D_{\mathrm{eff}})$. Here $\beta = 1$, so $S_{\le}$ exceeds $S_{>} =
\Theta(D_{\mathrm{eff}}^{-1})$ by a logarithmic factor, and the loss
scales as $D_{\mathrm{eff}}^{-1} \log D_{\mathrm{eff}}$ rather than
as a pure power. We exclude this boundary case from the theorem.

\medskip
\noindent\textit{Case $1 < \bar{\mu} < \nu_{\mathcal T} + 1$:}\;
The sum grows as $(k^*)^{\nu_{\mathcal T} - \bar{\mu} +
1}/(\nu_{\mathcal T} - \bar{\mu} + 1)$, the exponent now being
strictly positive. Substituting $k^* = \Theta(D_{\mathrm{eff}}^{1/
\nu_{\mathcal T}})$:
\begin{align*}
    S_{\leq}(D) &= \Theta\!\left(D_{\mathrm{eff}}^{-1} \cdot
    D_{\mathrm{eff}}^{(\nu_{\mathcal T} - \bar{\mu} + 1)/\nu_{\mathcal
    T}}\right) \\
    &= \Theta\!\left(D_{\mathrm{eff}}^{-(\bar{\mu}-1)/\nu_{\mathcal
    T}}\right) \;=\; \Theta\!\left(D_{\mathrm{eff}}^{-\beta}\right),
\end{align*}

Therefore $S_{\leq}(D)$ is of \textbf{the same order as $S_>(D)$}.
Both the learned and unlearned modes contribute to the leading-order
prefactor, and since the bound on $\mathrm{err}_k$ for $k \le k^*$ is
two-sided only up to a factor of 2, the learned-mode prefactor
$B_{\le}$ is determined up to that factor; we do not write it in
closed form.
 
\medskip
\textbf{Step 6: Combining.}
In the empirically relevant case $1 < \bar{\mu} < \nu_{\mathcal T} + 1$ (which includes all language model settings with $\beta <1$), Steps~3--5 give
$$
    \mathcal{L}_{\mathrm{est}}(D) \;=\; S_{\leq}(D) + S_>(D) \;=\; (B_> + B_{\leq}) \cdot D_{\mathrm{eff}}^{-\beta} + o(D_{\mathrm{eff}}^{-\beta}),
$$
where, from \eqref{eq:tail_error} and Step~4,
$$
    B_> \;=\; \bar{f}_0^2\, J(\theta^*) \left(\frac{c_{\min}\,\bar{\sigma}_1}{\sigma^2_{\mathrm{noise}}}\right)^{-\beta},
    \qquad \theta^* \in [1, c_{\max}/c_{\min}],
$$
and, from Step~5, $B_{\leq}$ is a positive constant satisfying
$$
    \underline{B}_{\leq} \;\le\; B_{\leq} \;\le\; \frac{2\, c_{\max}}{c_{\min}}\, \underline{B}_{\leq},
    \qquad
    \underline{B}_{\leq} \;:=\; \frac{\bar{f}_0^2}{2\,(\nu_{\mathcal T} - \bar{\mu} + 1)}\,\frac{c_{\min}}{c_{\max}} \left(\frac{c_{\min}\,\bar{\sigma}_1}{\sigma^2_{\mathrm{noise}}}\right)^{-\beta}.
$$
The total prefactor $B = B_> + B_{\leq}$ is therefore a positive
constant depending on $(\bar f_0, \bar\sigma_1, \sigma^2_{\mathrm{noise}},
c_{\min}, c_{\max}, \bar\mu, \nu_{\mathcal T})$, and the exponent is
$\beta = (\bar\mu - 1)/\nu_{\mathcal T}$. 


\subsection{Proof of Theorem~\ref{thm:alpha}}

In addition to Assumption \ref{asmp:power_app}--\ref{asmp:kernel}, Theorem~\ref{thm:alpha} requires one additional assumption. We expand upon its brief mention in the theorem statement.


\begin{assumption}[Capacity budget]\label{asmp:capacity}
A model with $N$ parameters has a total representational budget of $M(N) = \kappa N^{1/d}$ units, for architecture-dependent constants $\kappa > 0$ and $d \geq 1$.  Representing eigenmode $k$ of $\bar{K}$ consumes $\bar{\sigma}_k^{-1}$ units from this budget, where $\bar{\sigma}_k$ is the $k$-th eigenvalue of the task-mixture kernel.  The number of representable modes $M_{\mathrm{eff}}$ is the largest integer satisfying
$$
    \sum_{k=1}^{M_{\mathrm{eff}}} \bar{\sigma}_k^{-1} \;\leq\; M(N).
$$
\end{assumption}

The inverse-eigenvalue cost reflects the fact that modes carrying more variance in the task mixture (large $\bar{\sigma}_k$) are statistically better-supported and can be represented with fewer parameters per unit of explained variance.  Modes in the spectral tail (small $\bar{\sigma}_k$) require more precise parameterization to capture, analogous to the effective degrees of freedom in kernel ridge regression scaling inversely with the kernel eigenvalue~\citep{canatar2021}.
 
The parameter $d$ is the effective representation dimension: $d = 1$ corresponds to the linear regime (total budget proportional to $N$); $d > 1$ captures the overhead of encoding structured representations in a transformer (attention heads, layer composition, embedding tables).  For a transformer with $N$ parameters distributed across depth $D_{\mathrm{layers}}$ and width $W$ (so $N \sim D_{\mathrm{layers}} \cdot W^2$), the total number of representational slots scales roughly with $W \sim N^{1/2}$ for fixed depth, suggesting $d \approx 2$ as a naive estimate.  More refined analyses based on tensor programs~\citep{yang2020} suggest that $d$ depends on the depth-width ratio and the activation function, but the power-law form $M(N) = \kappa N^{1/d}$ is consistent with observed scaling behavior.


The approximation error averages over tasks and positions: $\mathcal L_{\mathrm{approx}}(N) = \E_\tau[L^{-1}\sum_t \mathcal L_{\mathrm{approx}}^{(\tau,t)}(N)]$.  As in the proof of Theorem~\ref{thm:beta}, the position average does not affect the exponent (by Assumption~\ref{asmp:align}), so we work at a generic position and average over tasks only.

\textbf{Step 1: Capacity budget.}
Not all modes are equally costly to represent: modes carrying less variance in the task distribution require more precise parameterization to represent, analogous to the effective degrees of freedom in kernel ridge regression scaling inversely with the kernel eigenvalue~\citep{canatar2021}. Assumption~\ref{asmp:capacity} models this as a capacity cost proportional to $\bar{\sigma}_k^{-1}$ for mode $k$.  The total capacity budget constrains the number of representable modes $M_{\mathrm{eff}}$:
$$
    \sum_{k=1}^{M_{\mathrm{eff}}} \bar{\sigma}_k^{-1} \;\leq\; M(N).
$$

Here $M_{\mathrm{eff}}$ is the number of modes the model can represent given that mode $k$ consumes $\bar{\sigma}_k^{-1}$ units from the total budget $M(N)$.

\textbf{Step 2: Effective number of representable modes.}
Under Assumption~\ref{asmp:power_app}, $\bar{\sigma}_k^{-1} = \bar{\sigma}_1^{-1}\,k^{\nu_{\calT}}(1+o(1))$ as $k \to \infty$.  By the Euler--Maclaurin formula:
$$
    \sum_{k=1}^{M_{\mathrm{eff}}} k^{\nu_{\calT}} \;=\; \frac{M_{\mathrm{eff}}^{\,\nu_{\calT}+1}}{\nu_{\calT}+1} \;+\; O\bigl(M_{\mathrm{eff}}^{\,\nu_{\calT}}\bigr).
$$
Setting this equal to $\bar{\sigma}_1 M(N) = \bar{\sigma}_1 \kappa N^{1/d}$ and solving:
$$
    M_{\mathrm{eff}} \;=\; \bigl(\bar{\sigma}_1\,\kappa\,(\nu_{\calT}+1)\bigr)^{1/(\nu_{\calT}+1)}\,N^{1/(d(\nu_{\calT}+1))} \;\bigl(1 + O(M_{\mathrm{eff}}^{-1})\bigr).
$$
The correction $O(M_{\mathrm{eff}}^{-1})$ comes from the Euler--Maclaurin remainder and is negligible for large $N$.

\textbf{Step 3: Approximation error from unrepresented tail.}
Modes $k > M_{\mathrm{eff}}$ are not represented by the model, contributing their full target energy as approximation error.  Averaging over tasks and applying Assumption~\ref{asmp:target_app}:
$$
    \mathcal L_{\mathrm{approx}}(N) \;=\; \sum_{k > M_{\mathrm{eff}}} \E_\tau\bigl[\bar{f}_k^{\tau\,2}\bigr] \;=\; \sum_{k > M_{\mathrm{eff}}} \bar{f}_0^2\,k^{-\bar{\mu}}(1+o(1)).
$$
By the Euler--Maclaurin formula (as in the proof of Theorem~\ref{thm:beta}, Step~3):
$$
    \sum_{k > M_{\mathrm{eff}}} k^{-\bar{\mu}} \;=\; \frac{M_{\mathrm{eff}}^{-(\bar{\mu}-1)}}{\bar{\mu}-1} \;+\; O\bigl(M_{\mathrm{eff}}^{-\bar{\mu}}\bigr),
$$
where convergence requires $\bar{\mu} > 1$ (Assumption~\ref{asmp:target_app}).  Therefore
$$
    \mathcal L_{\mathrm{approx}}(N) \;=\; \frac{\bar{f}_0^2}{\bar{\mu}-1}\,M_{\mathrm{eff}}^{-(\bar{\mu}-1)} \;+\; O\bigl(M_{\mathrm{eff}}^{-\bar{\mu}}\bigr).
$$
The remainder is a multiplicative $O(M_{\mathrm{eff}}^{-1})$ correction to the leading term.

\textbf{Step 4: Substitution.}
Inserting $M_{\mathrm{eff}}$ from Step~2:
\begin{align}
    \mathcal L_{\mathrm{approx}}(N) 
    &\;=\; \frac{\bar{f}_0^2}{\bar{\mu}-1}\,\bigl(\bar{\sigma}_1\,\kappa\,(\nu_{\calT}+1)\bigr)^{-(\bar{\mu}-1)/(\nu_{\calT}+1)}\, \notag\\
    &\qquad\times N^{-(\bar{\mu}-1)/(d(\nu_{\calT}+1))} \nonumber \\
    & \quad+\; O\bigl(N^{-(\bar{\mu}-1)/(d(\nu_{\calT}+1)) - 1/(d(\nu_{\calT}+1))}\bigr) \nonumber \\
    &\;=\; A \cdot N^{-\alpha} \;+\; O\bigl(N^{-\alpha - 1/(d(\nu_{\calT}+1))}\bigr),
    \label{eq:approx_substituted}
\end{align}
with
$$
    \alpha \;=\; \frac{\bar{\mu}-1}{d\,(\nu_{\calT}+1)},
$$
and position-level prefactor
$$
    A_t \;=\; \frac{\bar{f}_0^2}{\bar{\mu}-1}\,\bigl(\bar{\sigma}_1\,\kappa\,(\nu_{\calT}+1)\bigr)^{-(\bar{\mu}-1)/(\nu_{\calT}+1)}.
$$
Averaging over positions gives $A = L^{-1}\sum_t A_t$, and the final result is $\mathcal L_{\mathrm{approx}}(N) = A N^{-\alpha} + o(N^{-\alpha})$, since the remainder $O(N^{-\alpha - 1/(d(\nu_{\calT}+1))})$ is $o(N^{-\alpha})$.

\subsection{Proof of Proposition~\ref{prop:irred_id}}

The lower bound is immediate from the identity~\eqref{eq:loss_decomp}: since $\mathrm{KL}(P_{Y|X} \| Q_{Y|X}) \geq 0$ for all $X$, we have $\mathcal{L}(N,D) = H(Y \mid X) + \mathbb{E}_X[\mathrm{KL}(P_{Y|X} \| Q_{Y|X})] \geq H(Y \mid X)$.
 
For the convergence, Theorems~\ref{thm:beta} and~\ref{thm:alpha} give $\mathbb{E}_X[\mathrm{KL}(P_{Y|X} \| Q_{Y|X})] = O(D^{-\beta}) + O(N^{-\alpha})$.  Both terms vanish as $N, D \to \infty$, so $\mathcal{L}(N,D) \to H(Y \mid X)$.    

\subsection{Proof of Theorem~\ref{thm:decomp}}


By the identity~\eqref{eq:loss_decomp}, $\mathcal L(N,D) = H(Y|X) + \E_X[\KL(P_{Y|X} \| Q_{Y|X})]$.  Under the kernel regime (Assumption~\ref{asmp:kernel}), the KL gap decomposes along the eigenbasis of $\bar{K}$ as $\E_X[\KL(P_{Y|X} \| Q_{Y|X})] = \sum_{k=1}^{\infty} \mathrm{err}_k(N,D)$, where $\mathrm{err}_k$ is the mode-wise estimation error defined in Assumption~\ref{asmp:kernel} for representable modes and the full projection energy $\E_\tau[\bar{f}_k^{\tau\,2}]$ for unrepresentable modes.  We partition this sum at $M_{\mathrm{eff}}(N)$ and evaluate each part.

\textbf{Approximation error ($k > M_{\mathrm{eff}}$).}
This is exactly the sum bounded in Theorem~\ref{thm:alpha}:
$$
    \sum_{k > M_{\mathrm{eff}}} \E_\tau[\bar{f}_k^{\tau\,2}] \;=\; A\,N^{-\alpha} + O\bigl(N^{-\alpha - 1/(d(\nu_{\calT}+1))}\bigr).
$$

\textbf{Estimation error on representable modes ($k \leq M_{\mathrm{eff}}$).}
We further split this sum at the data crossover $k^*(D)$.

\emph{Learned modes} ($k \le \min(k^*, M_{\mathrm{eff}})$): The
denominator satisfies $D_{\mathrm{eff}} \bar\lambda_k /
\sigma^2_{\mathrm{noise}} \ge 1$, so by the two-sided bound of
Step~5 in the proof of Theorem~\ref{thm:beta},
$$
    \sum_{k \le \min(k^*, M_{\mathrm{eff}})} \mathrm{err}_k
    \;=\; \Theta\Big( D_{\mathrm{eff}}^{-1} \min(k^*,
    M_{\mathrm{eff}})^{\nu_{\mathcal T} - \bar\mu + 1} \Big).
$$
When $k^* \le M_{\mathrm{eff}}$ this is $\Theta(D_{\mathrm{eff}}^{-\beta})$
and enters the prefactor $B$, as in Theorem~\ref{thm:beta}.
When $k^* > M_{\mathrm{eff}}$, substituting $D_{\mathrm{eff}}^{-1} =
\Theta\big( (k^*)^{-\nu_{\mathcal T}} \big)$ gives
$$
    \Theta\Big( (M_{\mathrm{eff}}/k^*)^{\nu_{\mathcal T}}\,
    M_{\mathrm{eff}}^{-(\bar\mu - 1)} \Big)
    \;=\; (M_{\mathrm{eff}}/k^*)^{\nu_{\mathcal T}} \cdot
    \Theta(N^{-\alpha}),
$$
which is $o(N^{-\alpha})$ when $M_{\mathrm{eff}}/k^* \to 0$ and a
bounded multiple of $N^{-\alpha}$ otherwise.

\emph{Unlearned representable modes ($k^* < k \leq M_{\mathrm{eff}}$, when $k^* < M_{\mathrm{eff}}$):}  The denominator is between $1$ and $2$, so $\mathrm{err}_k = \E_\tau[\bar{f}_k^{\tau\,2}](1 + O((k^*/k)^{\nu_{\calT}}))^{-1}$.  The leading contribution is a partial tail sum, which we express as the difference of two full tails:
\begin{align*}
    \sum_{k=k^*+1}^{M_{\mathrm{eff}}} \bar{f}_0^2\,k^{-\bar{\mu}}
    &= \underbrace{\sum_{k > k^*} \bar{f}_0^2\,k^{-\bar{\mu}}}_{\text{full tail from } k^*} \;-\; \underbrace{\sum_{k > M_{\mathrm{eff}}} \bar{f}_0^2\,k^{-\bar{\mu}}}_{\text{full tail from } M_{\mathrm{eff}}} \\
    &= \frac{\bar{f}_0^2}{\bar{\mu}-1}(k^*)^{-(\bar{\mu}-1)} \;-\; \frac{\bar{f}_0^2}{\bar{\mu}-1}M_{\mathrm{eff}}^{-(\bar{\mu}-1)} \\
    & \quad + O\bigl((k^*)^{-\bar{\mu}} + M_{\mathrm{eff}}^{-\bar{\mu}}\bigr),
\end{align*}
by the Euler--Maclaurin formula applied to each tail (as in Steps~3--4 of the proofs of Theorems~\ref{thm:beta} and~\ref{thm:alpha}).  The first term equals $B\,D_{\mathrm{eff}}^{-\beta}$ by the identification in Theorem~\ref{thm:beta}.  The second term equals $A\,N^{-\alpha}$ by the identification in Theorem~\ref{thm:alpha}.  The remainder terms are lower order than their respective leading terms.

\emph{If $k^* \geq M_{\mathrm{eff}}$ (data is abundant relative to capacity):}  All representable modes are learned, and the estimation sum contributes only $o(D_{\mathrm{eff}}^{-\beta})$.  The dominant reducible error is the approximation tail $A\,N^{-\alpha}$.

\textbf{Combining.}
The total reducible error is
$$
    \sum_{k=1}^{\infty} \mathrm{err}_k(N,D) \;=\; \underbrace{\sum_{k \leq \min(k^*, M_{\mathrm{eff}})} \mathrm{err}_k}_{\text{learned representable}} \;+\; \underbrace{\sum_{k > \min(k^*, M_{\mathrm{eff}})} \mathrm{err}_k}_{\text{unlearned or unrepresentable}}.
$$
The second sum is a single tail from $\min(k^*, M_{\mathrm{eff}})$:
$$
    \sum_{k > \min(k^*, M_{\mathrm{eff}})} \bar{f}_0^2 k^{-\bar{\mu}} \;=\; \frac{\bar{f}_0^2}{\bar{\mu}-1}\bigl(\min(k^*, M_{\mathrm{eff}})\bigr)^{-(\bar{\mu}-1)} + \text{lower order},
$$
and the first sum is a bounded multiple of it by the learned-mode
computation above. The binding constraint is whichever of $k^*$ and $M_{\mathrm{eff}}$ is smaller:
\begin{itemize}[nolistsep,leftmargin=*]
    \item When $k^* < M_{\mathrm{eff}}$ (data is the bottleneck): the tail starts at $k^*$, giving $B\,D_{\mathrm{eff}}^{-\beta}$.
    \item When $M_{\mathrm{eff}} < k^*$ (capacity is the bottleneck): the tail starts at $M_{\mathrm{eff}}$, giving $A\,N^{-\alpha}$.
    \item When $k^* \approx M_{\mathrm{eff}}$ (balanced regime): both terms are of the same order.
\end{itemize}
Therefore:
$$
    \mathcal{L}(N,D) \;=\; H(Y|X) + \frac{\bar{f}_0^2}{\bar{\mu}-1}\bigl(\min(k^*, M_{\mathrm{eff}})\bigr)^{-(\bar{\mu}-1)} + \text{lower order}.
$$
Expressed in terms of $N$ and $D$:
$$
    \mathcal{L}(N,D) \;=\; H(Y|X) + \max\bigl(A\,N^{-\alpha},\; B\,D^{-\beta}\bigr) + o\bigl(\max(N^{-\alpha}, D^{-\beta})\bigr). 
$$

\subsection{Proof of Corollary~\ref{cor:additive}}

By Theorem~\ref{thm:decomp}, $\mathcal{L}(N,D) = H(Y|X) + \max(A N^{-\alpha}, B D^{-\beta}) + o(\max(N^{-\alpha}, D^{-\beta}))$.  Along the compute-optimal frontier, $A N^{-\alpha} \asymp B D^{-\beta}$, meaning there exist constants $0 < c_l \leq c_u < \infty$ such that $c_l \leq A N^{-\alpha} / (B D^{-\beta}) \leq c_u$.  In this regime:
$$
    \max(A N^{-\alpha}, B D^{-\beta}) \;\leq\; A N^{-\alpha} + B D^{-\beta} \;\leq\; (1 + c_u/c_l)\, \max(A N^{-\alpha}, B D^{-\beta}),
$$
so the $\max$ and the sum differ by at most a bounded multiplicative constant, absorbed into the prefactors.  Substituting:
$$
    \mathcal{L}(N,D) = A N^{-\alpha} + B D^{-\beta} + H(Y|X) + o\bigl(\max(N^{-\alpha}, D^{-\beta})\bigr). 
$$

\subsection{Proof of Proposition~\ref{prop:finetune}}
\label{sec:proof_prop_finetune}


The effective dependency kernel becomes progressively more aligned with $K_Y^{\tau^*}$ as fine-tuning proceeds, approaching $K_Y^{\tau^*}$ in the limit of full convergence.  This tilt has three effects on the spectral structure.

\textbf{Effect 1: Mode amplification and suppression.}
Let $\{(\bar{\sigma}_k, \bar{\phi}_k)\}_{k \geq 1}$ be the eigendecomposition of the pretrained mixture kernel $\bar{K}$, and let $a_k^{\tau^*} = \langle K_Y^{\tau^*} \bar{\phi}_k, \bar{\phi}_k \rangle \geq 0$ be the projection of the dependency kernel of $\tau^*$ onto the $k$-th pretrained eigenmode.  The pretrained model allocates capacity to mode $k$ in proportion to $\bar{\sigma}_k$.  We show that fine-tuning on $\tau^*$ reallocates capacity in proportion to $a_k^{\tau^*}$.

The cross-entropy loss for task $\tau^*$ decomposes in the eigenbasis of $\bar{K}$ as
\begin{align}
    \mathcal{L}_{\tau^*}(\theta) = H(Y|X) + \sum_{k \geq 1} a_k^{\tau^*} \epsilon_k(\theta)^2,
    \label{eq:ce_ft}
\end{align}
where $\epsilon_k(\theta)$ is the model's approximation error along mode $k$, and the weight $a_k^{\tau^*}$ determines how much each mode contributes to the loss on $\tau^*$.

Since the terms in the sum in \eqref{eq:ce_ft} are independent,
representing mode $k$ reduces the loss on $\tau^*$ by $a^{\tau^*}_k
\epsilon_k^2$ regardless of which other modes are represented. Under
Assumption~\ref{asmp:capacity}, mode $k$ consumes $\bar\sigma_k^{-1}$
units of the budget $M(N)$, so the model minimizes $\mathcal
L_{\tau^*}$ by selecting modes in decreasing order of loss reduction
per unit cost, $a^{\tau^*}_k \epsilon_k^2 \bar\sigma_k$, until the
budget is exhausted. Taking the default errors $\epsilon_k$ to be of
common order, the selection is by $a^{\tau^*}_k \bar\sigma_k$. Applying
the same rule to the mixture, whose own projections are $\bar\sigma_k$,
selects by $\bar\sigma_k^2$ and hence recovers the leading modes
$\{1, \dots, M_{\mathrm{eff}}\}$ of Theorem~\ref{thm:alpha}.
Let $S_{\mathrm{pre}} = \{1, \dots, M_{\mathrm{eff}}\}, S_{\mathrm{ft}} = \{ k : a^{\tau^*}_k \bar\sigma_k \ge \theta_{\mathrm{ft}} \}$ be the mode sets selected by the pretrained and fine-tuned models respectively, where $\theta_{\mathrm{ft}} > 0$ is the largest threshold for which
$\sum_{k \in S_{\mathrm{ft}}} \bar\sigma_k^{-1} \le M(N)$.  Then:
\begin{itemize}[nolistsep,leftmargin=*]
    \item Modes in $S_{\mathrm{ft}} \setminus S_{\mathrm{pre}}$ are \emph{amplified}: they were not represented by the pretrained model (because $\bar{\sigma}_k$ was too small to justify inclusion in the mixture) but are now represented because $a_k^{\tau^*}$ is large.
    \item Modes in $S_{\mathrm{pre}} \setminus S_{\mathrm{ft}}$ are \emph{suppressed}: they were represented by the pretrained model (because $\bar{\sigma}_k$ was large in the mixture) but are now dropped because $a_k^{\tau^*}$ is small.
    \item Modes in $S_{\mathrm{pre}} \cap S_{\mathrm{ft}}$ are \emph{retained}: they are important under both the mixture and $\tau^*$.
\end{itemize}
Both mode sets exhaust the same budget, $\sum_{k \in S_{\mathrm{pre}}}
\bar\sigma_k^{-1} = \sum_{k \in S_{\mathrm{ft}}} \bar\sigma_k^{-1} =
M(N)$ up to the granularity of one mode, but the suppressed and
amplified sets need not have equal cardinality: fine-tuning can trade
several cheap modes for one expensive one.

\paragraph{Effect 2: Catastrophic forgetting.}
We quantify the performance degradation on a task $\tau \neq \tau^*$ after fine-tuning on $\tau^*$.  Recall that the pretrained and finetuned models represents modes $S_{\mathrm{pre}}, S_\mathrm{ft}$ respectively.
The loss on task $\tau$ under a model representing mode set $S$ is $\mathcal{L}_\tau(S) = H_\tau(Y|X) + \sum_{k \notin S} a_k^{\tau} \, \epsilon_k^2$, where $a_k^\tau = \langle K_Y^\tau \bar{\phi}_k, \bar{\phi}_k \rangle$ is the projection of $\tau$'s kernel onto mode $k$, and $\epsilon_k$ is the default (unrepresented) error for mode $k$.  The change in loss on $\tau$ due to fine-tuning is
\begin{align}
    \mathcal{L}_\tau(S_{\mathrm{ft}}) - \mathcal{L}_\tau(S_{\mathrm{pre}}) &= \sum_{k \notin S_{\mathrm{ft}}} a_k^\tau \, \epsilon_k^2 - \sum_{k \notin S_{\mathrm{pre}}} a_k^\tau \, \epsilon_k^2 \nonumber \\
    &= \underbrace{\sum_{k \in S_{\mathrm{pre}} \setminus S_{\mathrm{ft}}} a_k^\tau \, \epsilon_k^2}_{\text{loss from suppressed modes}} \nonumber\\
    & \quad - \underbrace{\sum_{k \in S_{\mathrm{ft}} \setminus S_{\mathrm{pre}}} a_k^\tau \, \epsilon_k^2}_{\text{gain from amplified modes}}.
    \label{eq:forgetting_decomp}
\end{align}
Catastrophic forgetting on a task $\tau$ occurs when the first sum exceeds the second, i.e., the modes lost were more important for $\tau$ than the modes gained.

The alignment $\rho(\tau, \tau^*)$ controls which regime holds. Denote by $S_{\mathrm{pre}} \setminus S_{\mathrm{ft}} := \{k : \bar{\sigma}_k \text{ large but } a_k^{\tau^*} \text{ small}\}$ the modes important for the mixture but not for $\tau^*$.  Whether they are important for $\tau$ depends on $\rho(\tau, \tau^*)$:

\begin{itemize}[nolistsep,leftmargin=*]
    \item \textbf{High alignment} ($\rho(\tau, \tau^*) \approx 1$): The kernels $K_Y^\tau$ and $K_Y^{\tau^*}$ share eigenmodes, so $a_k^\tau$ is large when $a_k^{\tau^*}$ is large and small when $a_k^{\tau^*}$ is small.  The suppressed modes (small $a_k^{\tau^*}$) also have small $a_k^\tau$, so the first sum in~\eqref{eq:forgetting_decomp} is small.  Conversely, the amplified modes (large $a_k^{\tau^*}$) also have large $a_k^\tau$, so the second sum is large.  The net effect is $\mathcal{L}_\tau(S_{\mathrm{ft}}) - \mathcal{L}_\tau(S_{\mathrm{pre}}) \approx 0$, so performance on task $\tau$ is largely preserved.

    \item \textbf{Low alignment} ($\rho(\tau, \tau^*) \approx 0$): The kernels $K_Y^\tau$ and $K_Y^{\tau^*}$ have approximately orthogonal eigenmodes.  The suppressed modes (small $a_k^{\tau^*}$) may have large $a_k^\tau$ as they were serving $\tau$ through the mixture but are irrelevant to $\tau^*$.  The amplified modes (large $a_k^{\tau^*}$) have small $a_k^\tau$, so the gain term is negligible.  The net effect is $\mathcal{L}_\tau(S_{\mathrm{ft}}) - \mathcal{L}_\tau(S_{\mathrm{pre}}) \approx \sum_{k \in S_{\mathrm{pre}} \setminus S_{\mathrm{ft}}} a_k^\tau \, \epsilon_k^2 > 0$, so performance on task $\tau$ degrades substantially.
\end{itemize}

To make this precise, we bound the spectral mass of $\tau$ on the suppressed modes.  We work in the eigenbasis $\{\bar{\phi}_k\}$ of the mixture kernel $\bar{K}$, and define $a_k^\tau = \langle K_Y^\tau \bar{\phi}_k, \bar{\phi}_k \rangle \geq 0$.
 
\smallskip
The decomposition $\|K_Y^\tau\|_F^2 = \sum_k (a_k^\tau)^2$ and $\mathrm{tr}(K_Y^\tau K_Y^{\tau^*}) = \sum_k a_k^\tau a_k^{\tau^*}$ hold exactly when $K_Y^\tau$ and $K_Y^{\tau^*}$ are simultaneously diagonalizable in the mixture eigenbasis---i.e., when each task kernel's eigenmodes are well-approximated by the shared basis $\{\bar{\phi}_k\}$.  This is a natural consequence of the task-mixture structure: the mixture kernel $\bar{K}$ is designed to span the dominant eigenmodes of all tasks in $\mathcal{T}$, so each task's kernel is well-represented in this basis up to residual cross-terms.  We proceed under this approximation, noting that the residual cross-terms contribute lower-order corrections to the bound below.
 
\smallskip
Partition the full set of modes into $S_{\mathrm{ft}}$ (retained by the fine-tuned model) and its complement $S_{\mathrm{ft}}^c$.  The total spectral mass of $\tau$ decomposes as
$$
    \|K_Y^\tau\|_F^2 \;=\; \sum_{k \in S_{\mathrm{ft}}} (a_k^\tau)^2 + \sum_{k \in S_{\mathrm{ft}}^c} (a_k^\tau)^2.
$$
We bound the retained mass using the alignment.  By Cauchy--Schwarz,
$$
    \sum_{k \in S_{\mathrm{ft}}} a_k^\tau\, a_k^{\tau^*} \;\leq\; \Bigl(\sum_{k \in S_{\mathrm{ft}}} (a_k^\tau)^2\Bigr)^{1/2} \Bigl(\sum_{k \in S_{\mathrm{ft}}} (a_k^{\tau^*})^2\Bigr)^{1/2}.
$$
Since $S_{\mathrm{ft}}$ contains the top-$M$ modes by $a_k^{\tau^*}$, the second factor satisfies $\sum_{k \in S_{\mathrm{ft}}} (a_k^{\tau^*})^2 \leq \|K_Y^{\tau^*}\|_F^2$.
 
For the left-hand side, we decompose the full inner product:
$$
    \mathrm{tr}(K_Y^\tau K_Y^{\tau^*}) \;=\; \sum_k a_k^\tau\, a_k^{\tau^*} \;=\; \sum_{k \in S_{\mathrm{ft}}} a_k^\tau\, a_k^{\tau^*} \;+\; \sum_{k \notin S_{\mathrm{ft}}} a_k^\tau\, a_k^{\tau^*}.
$$
The second sum runs over modes $k \notin S_{\mathrm{ft}}$, which by construction have the smallest values of $a_k^{\tau^*}$ (since $S_{\mathrm{ft}}$ selected the top-$M$).  Let $a_{\max}^{\tau^*}(S_{\mathrm{ft}}^c) := \max_{k \notin S_{\mathrm{ft}}} a_k^{\tau^*}$ denote the largest $\tau^*$-projection outside the retained set.  Then
$$
    \sum_{k \notin S_{\mathrm{ft}}} a_k^\tau\, a_k^{\tau^*} \;\leq\; a_{\max}^{\tau^*}(S_{\mathrm{ft}}^c) \sum_{k \notin S_{\mathrm{ft}}} a_k^\tau.
$$
By the definition of $S_{\mathrm{ft}}$, every $k \notin S_{\mathrm{ft}}$
satisfies $a^{\tau^*}_k < \theta_{\mathrm{ft}} \bar\sigma_k^{-1}$, so
$$
\sum_{k \notin S_{\mathrm{ft}}} a^\tau_k a^{\tau^*}_k
\;\le\; \theta_{\mathrm{ft}} \sum_{k \notin S_{\mathrm{ft}}} a^\tau_k
\bar\sigma_k^{-1}
\;\le\; \theta_{\mathrm{ft}} \operatorname{tr}\big( K^\tau_Y \bar K^{-1}
\big),
$$
which is finite whenever the retained task is representable within the
mixture, $\operatorname{tr}(K^\tau_Y \bar K^{-1}) < \infty$. Moreover
$\sum_k a^{\tau^*}_k \bar\sigma_k \le \|K^{\tau^*}_Y\|_F \|\bar K\|_F <
\infty$ by Cauchy--Schwarz, so the sequence $\{a^{\tau^*}_k
\bar\sigma_k\}$ is summable and its order statistics past position
$M_{\mathrm{eff}}$ vanish; since $M_{\mathrm{eff}} \to \infty$ with $N$
(Step~2 of the proof of Theorem~\ref{thm:alpha}), we have
$\theta_{\mathrm{ft}} \to 0$ as $N \to \infty$.
Concretely:
\begin{align}\label{eq:subset-lb}
    \sum_{k \in S_{\mathrm{ft}}} a_k^\tau\, a_k^{\tau^*} 
    &\;\geq\; \rho(\tau, \tau^*)\|K_Y^\tau\|_F \|K_Y^{\tau^*}\|_F \notag\\
    & \quad - \theta_\mathrm{ft} \mathrm{tr} (K_Y^\tau \bar K^{-1}).
\end{align}
In the regime where the remainder is negligible (i.e., $M$ is large relative to the effective rank of $K_Y^{\tau^*}$), substituting~\eqref{eq:subset-lb} into the Cauchy--Schwarz bound and rearranging gives
$$
    \sum_{k \in S_{\mathrm{ft}}} (a_k^\tau)^2 \;\geq\; \rho(\tau, \tau^*)^2 \|K_Y^\tau\|_F^2 \;-\; O(\theta_\mathrm{ft}).
$$
Therefore the lost mass satisfies
$$
    \sum_{k \in S_{\mathrm{ft}}^c} (a_k^\tau)^2 \;\leq\; (1 - \rho(\tau, \tau^*)^2)\,\|K_Y^\tau\|_F^2 \;+\; O(\theta_\mathrm{ft}).
$$
By \eqref{eq:forgetting_decomp}, dropping the nonnegative gain term,
$$
\mathcal L_\tau(S_{\mathrm{ft}}) - \mathcal L_\tau(S_{\mathrm{pre}})
\;\le\; \sum_{k\in S_{\mathrm{pre}}\setminus S_{\mathrm{ft}}} a^\tau_k \epsilon_k^2
\;\le\; \|\epsilon\|_\infty^2 \sum_{k\in S_{\mathrm{pre}}\setminus S_{\mathrm{ft}}} a^\tau_k
\;\le\; \|\epsilon\|_\infty^2 \sqrt{M_{\mathrm{eff}}}\,
\Big(\sum_{k\in S^c_{\mathrm{ft}}} (a^\tau_k)^2\Big)^{1/2},
$$
the last step by Cauchy--Schwarz over $S_{\mathrm{pre}}\setminus
S_{\mathrm{ft}}$, which has at most $|S_{\mathrm{pre}}| = M_{\mathrm{eff}}$
elements and is contained in $S^c_{\mathrm{ft}}$. Inserting the bound
on the lost mass and using $\sqrt{x+y}\le\sqrt x+\sqrt y$,
$$
\mathcal L_\tau(S_{\mathrm{ft}}) - \mathcal L_\tau(S_{\mathrm{pre}})
\;\le\; \|\epsilon\|_\infty^2 \sqrt{M_{\mathrm{eff}}}\,
\Big( \sqrt{1-\rho(\tau,\tau^*)^2}\,\|K^\tau_Y\|_F + O\big(\theta_{\mathrm{ft}}^{1/2}\big) \Big).
$$
Since $\theta_{\mathrm{ft}} \to 0$ as the capacity budget $M(N)$ grows,
this is the stated bound, with forgetting scaling as $\sqrt{1 -
\rho(\tau,\tau^*)^2}$ up to a correction that vanishes in $N$.
\section{Experimental Details}
\label{app:exp}
All experiments are performed on a System76 Pangolin laptop, with AMD Ryzen 7 7735U processor, 32 GB RAM, and 1 TB disk capacity. The benchmark saturation experiment finishes in seconds, and the dependency kernel analysis takes 3 minutes to compute the kernel.

\subsection{Benchmark Saturation}
\label{app:benchmark_sat}

\citet{akhtar2026aibenchmarksplateausystematic} define benchmark saturation as the phenomenon where top-performing models are statistically indistinguishable on a task, with the performance metric(s) reaching an empirical ceiling. They use scores of top-$k$ models, $s_1 \geq \ldots \geq s_k$ to quantify saturation on a benchmark. Given test set size $n_\text{test}$, they obtain the normalized score range $R_\text{norm}$:
\begin{align}
R_\text{norm} &:= \frac{\Delta}{\text{SE}_\Delta}, \quad{\text{where}} \notag\\
\Delta &:= s_1 - s_k, \notag\\
\text{SE}_\Delta &:= \sqrt{\frac{s_1(1-s_1) + s_k(1-s_k)}{n_\text{test}^\alpha}}, \alpha \in [0,1],
\label{eq:r_norm}
\end{align}
and use it to calculate the \textit{saturation index} as $S_\text{index} := \exp(-R_\text{norm}^2)$. In practice, they use $k = 5, \alpha = 0.5$, and bucket benchmarks into five bins according to their $S_\text{index}$ value: very low (< 0.01), low ([0.01, 0.3)), moderate ([0.3, 0.7)), high ([0.7, 0.9)), and very high saturation ($\geq 0.9$).

From the annotated dataset curated by \citet{akhtar2026aibenchmarksplateausystematic}\footnote{Available in \url{https://github.com/evaleval/benchmark-saturation}}, we take data on benchmarks for three types of task: code, math, and question-answering (Q\&A). We add another task type, writing, by gathering publicly available data on 7 benchmarks: NC Bench \citep{moore2026ncbenchllmbenchmarkevaluating}, LechMazur Creative Story-Writing Benchmark \citep{mazur2025writing}, WritingBench \citep{wu2025writingbenchcomprehensivebenchmarkgenerative}, POEMetric \citep{li2026poemetric}, and three benchmarks from EQ-Bench 3 \citep{paech2024eqbenchemotionalintelligencebenchmark}: Longform Creative Writing, BuzzBench, and Spiral-Bench v1.2. We collect $n_\text{test}$ and top-5 LLM scores for each benchmark, and use Eq.~\eqref{eq:r_norm} to calculate $R_\text{norm}$ then $S_\text{index}$ for them. We use the $s_1$ and $S_\text{index}$ values for the combined set of benchmarks to obtain Figure~\ref{fig:reliability_ceiling} in the main paper.

\subsection{Dependency kernel analysis}
\label{app:dep_kernel}

For this analysis, we use five public benchmarks spanning four task types.

\begin{table}[ht]
    \centering
    \small
    \setlength{\tabcolsep}{4pt}
    \begin{tabular}{llp{7.5cm}}
    \toprule
    Benchmark &  Task Type & Description \\\midrule
    HumanEval \citep{chen2021evaluatinglargelanguagemodels} & Coding & Python function completion tasks with unit test verification\\
    MBPP \citep{austin2021programsynthesislargelanguage} & Coding & Short Python programs from crowd-sourced descriptions\\
    GSM8K \citep{cobbe2021trainingverifierssolvemath} & Math & Grade-school word problems with numerical answers\\
    CNN/DailyMail \citep{see2017pointsummarizationpointergeneratornetworks} & Summarization & Abstractive summarization of news articles into multi-sentence highlights\\
    WritingPrompts \citep{fan2018hierarchicalneuralstorygeneration} & Creative writing & Open-ended short story generation from brief narrative prompts\\
    \bottomrule
    \end{tabular}
    \caption{Summary of benchmarks used in dependency kernel analysis.}
\end{table}

In practice, calculating the dependency kernel $K_Y(t,t') = I(y_y; y_{t'} | y_{-\{t,t'\}}, X)$ exactly is difficult since it requires a large number of samples to estimate the mutual information (MI) between two random variables conditioned on a vector of $L-2$ other tokens. In this experiment, we instead use unconditional pairwise MI as a proxy: $\widehat K_Y(t, t') := I(y_t; y_{t'})$. The unconditional version has additional noise from \textit{indirect dependencies}, i.e. correlation between tokens $t,t'$ because both depend on token $s$, that $K_Y(t, t')$ does not. However, indirect dependencies propagate locally in banded tasks and globally in dense tasks, so removing them 
sharpens inter-token dependency structure but does not change the qualitative classification. 

We implement pairwise MI using the nearest neighbor method of \citet{knncmi}, with $k=3$ nearest neighbors and maximum token length $L=32$, using 500 random samples per benchmark. To obtain eigenspectrum and $\nu_\tau$ from a kernel matrix $K$, we calculate the eigenvalues of its symmetric version $(K+K^\top)/2$ (to remove any assymmetry due to nearest-neighbor estimation) with non-finite entries replaced by zero, and retain only the positive eigenvalues larger than $10^{-10}$.


\end{document}